\documentclass[10pt, conference, letterpaper]{IEEEtran}

\usepackage{cuted}
\usepackage{epsfig}
\usepackage{amsmath}
\usepackage{amsthm}
\usepackage{color}
\usepackage{flushend}
\usepackage{enumitem}
\setlist[itemize]{itemsep=0pt,parsep=0pt,topsep=-1mm}

\usepackage{subfigure}
\usepackage{graphicx}
\usepackage{cite}
\usepackage{amssymb}
\usepackage{fixmath}
\usepackage{caption}

\usepackage{algorithm2e}
 \usepackage{color}
\usepackage{setspace}
\usepackage{fontenc}
\usepackage[final]{changes}
\usepackage{soul}
\usepackage{xcolor} 

\newcommand{\bm}[1]{\mbox{\boldmath{$#1$}}}

\newtheorem{corollary}{Corollary}

\SetKwInOut{Input}{Input}
\SetKwInOut{Output}{Output}

\begin{document}

\IEEEoverridecommandlockouts

\title{Deep Reinforcement Learning-Based User Association in Hybrid LiFi/WiFi Indoor Networks}

\author{Peijun Hou, Nan Cen\\
Department of Computer Science,
Saint Louis University, St. Louis, MO 63103\\
Email:{\{peijun.hou, nan.cen\}@slu.edu}
}

\maketitle
\pagestyle{empty}

\begin{abstract}
Hybrid light fidelity (LiFi) and wireless fidelity (WiFi) indoor networks has been envisioned as a promising technology to alleviate radio frequency spectrum crunch to accommodate the ever-increasing data rate demand in indoor scenarios. The hybrid LiFi/WiFi indoor networks can leverage the advantages of fast data transmission from LiFi and wider coverage of WiFi, thus complementing well with each other and further improving the network performance compared with the standalone networks. However, to leverage the co-existence, several challenges should be addressed, including but not limited to user association, mobility support, and efficient resource allocation. Therefore, the objective of the paper is to design a new user-access point association algorithm to maximize the sum throughput of the hybrid networks. We first mathematically formulate the sum data rate maximization problem by determining the AP selection for each user in indoor networks with consideration of user mobility and practical capacity limitations, which is a nonconvex binary integer programming problem. To solve this problem, we then propose a sequential-proximal policy optimization (S-PPO) based deep reinforcement learning method. Extensive simulations are conducted to evaluate the proposed method by comparing it with exhaustive search (ES), signal strength strategy (SSS), and trust region policy optimization (TRPO) methods. Comprehensive simulation results demonstrate that our solution algorithm can outperform SSS by about 32.25\% of the sum throughput and 19.09\% of the fairness on average, and outperform TRPO by about 10.34\% and 10.23\%, respectively.

\end{abstract}


\IEEEpeerreviewmaketitle
\section{Introduction}\label{sec:intro}
Over the past few years, the usage of the internet has been continuously increasing. According to the latest data, people spend an average of 6 hours and 58 minutes daily on screens connected to the internet \cite{ScreenTimeStats2023}. Moreover, an increasing number of applications require high-speed support, such as video calls, VR gaming, streaming media, and so on. However, we are facing a global digital divide, i.e., internet speeds in urban areas are often much faster than in rural areas, due to the generally less developed internet infrastructure in rural locations. Visible light communication (VLC), where light-emitting diodes (LEDs) can be used to transmit data by optical spectrum, has been envisioned as a promising solution for last-mile access because of its high bandwidth, enhanced security, electromagnetic interference-free nature, and easy integration with existing infrastructure \cite{komine2004fundamental, cen2019lanet, my, 10060808, 10225783, 8574917}. In recent years, light fidelity (LiFi) has been proposed, which utilizes VLC for data transmission, and can be installed on existing lighting infrastructures, thus realizing a dual-purpose system that provides illumination and communication simultaneously.

Unlike wireless fidelity (WiFi) technology, operating on the crowded radio frequency (RF) spectrum, LiFi utilizes a large portion of the unregulated spectrum ranging from 400 to 800 THz. The main advantages of LiFi technology include: (i) high-speed data transmission potential relying on the spectrum of orders of magnitude \cite{6967750}; (ii) safe in RF-restricted environments because of no electromagnetic interference to RF, such as aircraft and hospital; (iii) inherently secure due to the low-penetration property \cite{cen2019lanet}.
Though LiFi offers significant advantages, it also has limitations: (i) limited field of view (FoV), usually a few meters for a single AP; (ii) sensitive to obstacles due to low penetration \cite{9351549}.

Integrating LiFi with existing WiFi networks to form hybrid LiFi/WiFi networks provides a solution that leverages the advantages of fast data transmission of LiFi and wider coverage range of WiFi, thus complementing well with each other to provide a seamless, high data-rate transmission in indoor networks \cite{rahaim2011hybrid}.
Recently, a number of studies have been conducted with the objective of improving the spectrum efficiency of hybrid LiFi/WiFi networks. 
The authors in \cite{8926487} propose a Q-learning based dual-timescale power allocation approach for multi-homing hybrid RF/VLC networks, aiming at meeting the quality of service (QoS) demands of users. A learning-based solution for optimizing network selection, subchannel distribution, and power allocation is proposed in \cite{8792078} to satisfy the diverse demands of users in heterogeneous RF/VLC networks. The work in \cite{9424627} explores the topology control problem in a hybrid VLC/RF vehicular ad-hoc network, and proposes a distributed algorithm.
A fuzzy logic based dynamic handover scheme for indoor LiFi and RF hybrid networks is proposed in\cite{7510823} to reduce the handover overhead. Among these efforts, network selection plays a crucial role in enhancing spectrum efficiency. However, traditional user-access point (AP) selection methods for homogeneous networks, such as signal strength strategy (SSS) \cite{pavon2003link} are not applicable as they will make the WiFi APs prone to overloading\footnote{In hybrid networks, WiFi has a larger coverage area but lower capacity compared with LiFi, thus WiFi AP will be likely to overload when SSS based approaches are adopted.}. Moreover, the dynamic characteristics of user requirements and mobility in indoor heterogeneous networks make the user-access point association problem more challenging.

In recent years, a few studies \cite{basnayaka2017design, wu2019mobility, 8123892, 8374416, ahmad2020reinforcement, 8943127, 10661225, 7876858, wu2017access, 10367833} have been conducted to propose
user-AP association methods specifically for the hybrid LiFi/WiFi indoor networks. 
However, most traditional methods, such as game theory \cite{10661225, 7876858}, and fuzzy logic-based AP selection \cite{wu2017access}, do not consider user mobility, which is critical for real-world applications where users frequently move and experience dynamic connectivity. Moreover, conventional methods often assume that channel state information (CSI) is perfectly known. However, in practical applications, CSI is typically unpredictable and susceptible to interference. To effectively address the dynamic and complex decision-making challenges posed by user mobility in wireless environments, adopting learning-based methods emerges as an effective solution \cite{8374416, ahmad2020reinforcement, 10367833}. Recently proposed supervised learning methods \cite{10367833} require a large set of training datasets, which is difficult to obtain in practical wireless networks. In contrast, deep reinforcement learning (DRL) can learn optimal behavior through continuous trial and error by interacting with the time-varying networking environments without a strict requirement of collecting a large volume of training datasets, thus providing a more efficient approach to optimize user association that adapts as the network changes.
Yet, current DRL works in \cite{8374416, ahmad2020reinforcement} do not consider the practical capacity limitations of APs, which makes the scenario unrealistic. 

To bridge the gap, our objective is to propose a robust user-AP association method for indoor hybrid LiFi/WiFi networks with either static or mobile users, aiming to maximize the achievable sum data rate. Due to the binary nature of the user-AP association variables, the resulting formulated problem is classified as a nonconvex binary integer programming (BIP) problem, typically NP-hard, implying that finding a globally optimal solution within polynomial time is unfeasible. Therefore, we propose a sequential-proximal policy optimization (S-PPO) based solution method to determine the user-AP association. The main contributions of this paper are outlined as follows:

\begin{itemize}
\item \emph{User-AP association formulation.} We mathematically formulate the user-AP association problem for indoor hybrid LiFi/WiFi networks with the objective of maximizing the sum throughput by jointly considering the network capacity limitation and the user mobility patterns. 
\item \emph{Sequential-proximal policy optimization based solution algorithm.} To solve the formulated binary integer programming problem, we propose a solution algorithm based on proximal policy optimization (PPO), an advanced reinforcement learning framework that combines the benefits of both actor-critic and policy gradient methods. 
Moreover, we design an action space decomposition strategy that significantly reduces the action space by decomposing the entire association policy into a sequencing policy. 
\item \emph{Performance evaluation.} We evaluate the performance of the proposed S-PPO method with respect to optimality, convergence, robustness, computational complexity, scalability, and fairness. We also compare the proposed S-PPO with exhaustive search (ES), signal strength strategy, and trust region policy optimization (TRPO) methods through extensive simulation experiments. Comprehensive simulation results show that the proposed S-PPO outperforms the other methods in terms of sum data rate, fairness, and runtime under different network topologies and capacity limitations. 

\end{itemize}

The remainder of the paper is organized as follows. 
Section~\ref{sec:relatedwork} reviews state-of-the-art. 
In Section~\ref{sec:sys}, we introduce the system model for the hybrid LiFi/WiFi networks, user-AP association model, and random waypoint model. The centralized optimization problem formulation is present in Section~\ref{sec:problem}. Subsequently, we discuss the proposed S-PPO method in Section~\ref{sec:solu}. Lastly, Section~\ref{sec:performance} evaluates the performance of the proposed method, followed by an analysis of limitations and future direction in Section~\ref{sec:discussion}, with conclusions highlighted in Section~\ref{sec:conclusion}.

\section{Related Work}\label{sec:relatedwork}
\subsection{Optimization Based Methods} \label{sec:sec:tradi}
There are some efforts that have been made in the literature to solve the user-AP association problem in hybrid LiFi/WiFi networks \cite{basnayaka2017design, wu2017access, guerin2021towards, ahmad2020reinforcement, 8374416}. In \cite{wu2017access}, a two-stage fuzzy logic based system is proposed to assign users to APs. The proposed system calculates the accessibility scores of users, and then assigns users to WiFi APs in descending order of scores until the maximum capacity of WiFi APs is reached. The remaining users are then served by LiFi APs. With this approach, the throughput of the system has been greatly improved. The authors in \cite{basnayaka2017design} design a data rate threshold in hybrid LiFi/WiFi networks, all the user equipment first connect to LiFi APs, and those below the threshold switch to WiFi APs. The simulation results show that the hybrid networks have higher spectral efficiency than the stand-alone network. The authors in \cite{guerin2021towards} propose a solution that utilizes the individual throughput of each station and overall network energy consumption to determine the association between stations and APs. Compared with baseline methods, the proposed approach achieves better trade-offs in terms of throughput and energy efficiency.

However, these studies assume that users remain stationary, which does not align with the dynamics of user behavior in the real world. Furthermore, the interference among LiFi APs is not considered in \cite{guerin2021towards}, which makes it lack generality.
Different from previous works, we take into account user mobility and inter-cell interference for LiFi APs. As users move, channel information changes, thereby complicating the user-AP association problem. 

\subsection{Reinforcement Learning Based Methods} \label{sec:sec:learn}
Reinforcement learning (RL) based methods have attracted considerable attention from researchers for their potential to solve the user-AP association problem in hybrid LiFi/WiFi networks. In \cite{8374416}, a Q-learning with knowledge transfer method is proposed to enable online networking selection, which 
outperforms the traditional Q-learning model and has a faster convergence speed. However, the method relies on knowledge or contextual information, which may vary across different network scenarios, and sometimes the contextual information is not available.
The work in \cite{ahmad2020reinforcement} implements trust region policy optimization to provide an optimal user association strategy with consideration of different user mobility patterns.
The results show that the proposed RL method outperforms the iterative algorithm and signal strength strategy method in most cases. In \cite{9468898}, a TRPO-based method is also proposed for load balancing in heterogeneous LiFi-WiFi networks. However, the TRPO based method used in this work is inefficient in terms of computational expenses and sample efficiency due to second-order optimization. 

Inspired by the above observations, we proposed a fast converging and computationally efficient sequential proximal policy optimization based actor-critic reinforcement learning algorithm to tackle the user-AP association problem in hybrid LiFi/WiFi indoor networks, taking into account the mobility of users as well as network capacity constraints. 

\section{System Model}\label{sec:sys}

\begin{figure}[t]
\begin{center}
\includegraphics[width=0.3\textwidth]{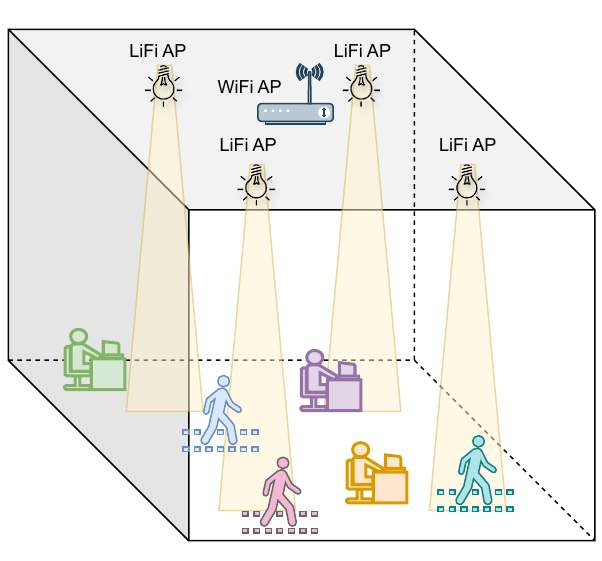}
\caption{\small{Indoor hybrid LiFi/WiFi networks.}} 
\vspace{-6mm}
\label{fig:real}
\end{center}
\end{figure}

We consider an indoor hybrid LiFi/WiFi network as illustrated in Fig. \ref{fig:real}. The indoor scenario consists of a set of $\mathcal{L}$ = \{1, 2, ..., $l$, ..., $L$\} LiFi APs and a set of $\mathcal{W}$ = \{1, 2, ..., $w$, ..., $W$\} WiFi APs. We assume that there is a set of $\mathcal{K}$ = \{1, 2, ..., $k$, ..., $K$\} active users uniformly distributed over the room, where users can only be served by either one LiFi AP or one WiFi AP. We assume each AP adopts time division multiple access (TDMA) to control the channel access for multiple users with predefined time slots. 
From a practical perspective, driven by the finite capacity limitation of the networks, we also limit the maximum number of connected users to each LiFi AP and WiFi AP as $N_{l}$ and $N_{w}$ as in \cite{9127161}.

\subsection{LiFi Channel Model} \label{sec:sec:lifi}

\begin{figure}[t]
\begin{center}
\includegraphics[width=0.319\textwidth]{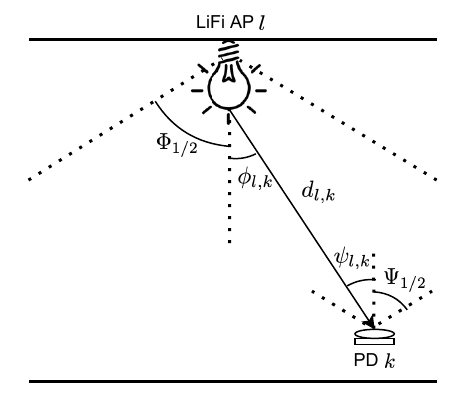}
\caption{\small{LiFi LoS Channel Model.}} 
\vspace{-6mm}
\label{fig:LOS}
\end{center}
\end{figure}

\normalsize {In this study, we consider the line of sight (LoS) downlink path in LiFi networks. Our following analysis can be easily extended to scenarios with LoS and non-LoS paths. The geometry of the LiFi LoS propagation is shown in Fig. \ref{fig:LOS}, where the LED transmitter -- LiFi AP $l$ is installed on the ceiling, directly facing downwards, and the receiving device photodiode (PD) is installed on each user $k$, facing directly upwards. 
$d_{l,k}(t)$, $\psi_{l,k}(t)$, and $\phi_{l,k}(t)$ are the distance between the LiFi AP $l$ and the user $k$, the angle of incidence, and the angle of irradiation at time slot $t$, respectively. $\rm\Psi_{1/2}$ is the half angle of the PD's FoV, and $\rm\Phi_{1/2}$ is the half-intensity radiation angle of LiFi AP. At time slot $t$, if user $k$ is served by LiFi AP $l$, the optical channel gain of the LoS channel can then be given as \cite{kahn1997wireless}: }

\setlength{\abovedisplayskip}{1pt}

\begin{align} \label{eq:channelgain}
&H^{lf}_{l,k}(t) = \notag \\ 
&\begin{cases} 
      \frac{(m+1)A}{2\pi d_{l,k}^2(t)} \cos^{m}(\phi_{l,k}(t))T_s(\psi_{l,k})g(\psi_{l,k}(t))\cos(\psi_{l,k}(t))\; 
       \\ \qquad\qquad\qquad\qquad\qquad\qquad\qquad\,\, \;  0 \le \psi_{l,k}(t) \le \rm\Psi_{1/2}, \\
      0\; \qquad\qquad\qquad\qquad\qquad\qquad \qquad  \text{otherwise} ,
      \end{cases}
\end{align}
where $m$ 
is the Lambertian index and is given by the semi-angle $\rm\Psi_{1/2}$ at half illuminance power of the LED as $m = \rm-ln2/{\ln(\cos \Psi_{1/2}})$. 
$A$ is the physical area of the PD.
$T_s(\psi_{l,k})$ denotes the gain of an optical filter. $g(\psi_{l,k}(t))$ is the optical concentrator gain
at time slot $t$, given as: 


\begin{align} \label{eq:concentratorgain}
g(\psi_{l,k}(t)) =\begin{cases}
       \frac{n^2}{\sin^2\rm\Psi_{1/2}} \; & 0 \le \psi_{l,k}(t) \le \rm\Psi_{1/2},  \\
       0 \; &  \text{otherwise},
      \end{cases}
\end{align}where $n$ is the refractive index. When user $k$ is connected to LiFi AP $l$ at time slot $t$, the signal-to-interference-plus-noise ratio (SINR) is calculated as:


\begin{align} \label{eq:lfsinr}
\gamma_{l,k}^{lf}(t) =
      \frac{(\kappa H^{lf}_{l,k}(t)P_{lf})^2}{\mathcal{N}_{lf}B_{lf}+\underset{j\neq l}{\sum} (\kappa H_{j,k}^{lf}(t)P_{lf})^2} ,
\end{align}where $\kappa$ denotes the optical to electric conversion efficiency of the adopted PD; $H_{l,k}^{lf}(t)$ is the channel gain between the user $k$ and the connected LiFi AP $l$ at time slot $t$; $P_{lf}$ and $B_{lf}$ are the transmitted optical power and modulation bandwidth of LiFi APs, respectively; $\mathcal{N}_{lf}$ represents the power spectral density of LiFi noise; $H_{j,k}^{lf}(t)$ is the channel gain between the user $k$ and the interfering LiFi AP $j$ at time slot $t$. 
Since the signals in LiFi are non-negative, Shannon capacity cannot be used directly. Instead, we use the lower bound in \cite{6636053}, the achievable data rate between LiFi AP $l$ and user $k$ at time slot $t$ can then be given as:

\begin{align} \label{eq:lfdatarate}
r_{l,k}^{lf}(t) = 
      \frac{B_{lf}}{2}\text{log}_{2}(1+\frac{e}{2\pi}\gamma_{l,k}^{lf}(t)).
\end{align}

\subsection{WiFi Channel Model} \label{sec:sec:wifi}
For WiFi networks, 
the path loss model defined in \cite{perahia2013next} is adopted, where free space loss and shadow fading loss due to large-scale obstructions are considered, given as: 

\begin{align} \label{eq:pathloss}
L(d_{w,k}(t)) =\begin{cases}
    \ \ L_{FS}(d_{w,k}(t))+X_{SF} \; & d_{w,k}(t) \le d_{BP},  \\
    \begin{array}{l}L_{FS}(d_{w,k}(t))+X_{SF}\\+35\text{log}_{10}(\frac{d_{w,k}(t)}{d_{BP}})\end{array} \; & d_{w,k}(t) > d_{BP},
     \end{cases}
\end{align}where $d_{w,k}(t)$ is the distance between the WiFi AP $w$ and the user $k$ at time slot $t$; $d_{BP}$ is the breakpoint distance; $X_{SF}$ denotes the shadow fading loss, which is a Gaussian random variable with zero mean and 3 dB deviation if the distance is smaller than or equal to the breakpoint, and 5 dB deviation otherwise; the free space loss $L_{FS}$ is calculated as:

\begin{align} \label{eq:freeloss}
L_{FS}(d_{w,k}(t)) =
      20\text{log}_{10}(d_{w,k}(t))+20\text{log}_{10}(f_c)-147.5, \; 
\end{align}with $f_c$ denoting the central carrier frequency. At time slot $t$, the channel transfer function between user $k$ and WiFi AP $w$ can be expressed as:

\begin{align} \label{eq:channeltransfer}
H_{w,k}^{wf}(t) =
      \sqrt{10^{\frac{-L(d_{w,k}(t))}{10}}}(\sqrt{\frac{K}{1+K}}e^{j\phi}+\sqrt{\frac{1}{1+K}}X), \; 
\end{align}with $K$ representing the Rician $K$-factor \cite{perahia2013next}. For the LoS path, $K=1$ if $d_{w,k}(t) \le d_{BP}$, otherwise $K= 0$. 
$\phi$ is the angle of arrival or departure of the LoS component. $X$ is a complex Gaussian random variable with zero mean and unit variance. Then, the SINR $\gamma_{w,k}^{wf}(t)$ for the WiFi channel model at time slot $t$ is given as:

\begin{align} \label{eq:wfsinr}
\gamma_{w,k}^{wf}(t) =
      \frac{|H_{w,k}^{wf}(t)|^2P_{wf}}{\mathcal{N}_{wf}B_{wf}}, \; 
\end{align}where $\mathcal{N}_{wf}$ represents the power spectral density of noise of the WiFi networks; $P_{wf}$ and $B_{wf}$ are the transmitted power and modulation bandwidth of each WiFi AP, respectively.
The achievable data rate $r_{w,k}^{wf}(t)$ between the WiFi AP $w$ and user $k$ can be calculated using:

\begin{align} \label{eq:wfdatarate}
r_{w,k}^{wf}(t) = 
      \frac{B_{wf}}{2}\text{log}_{2}(1+\gamma_{w,k}^{wf}(t)).
\end{align}

\subsection{User-AP Association Model} \label{sec:sec:userass}
We consider single-home accommodation for users in favor of tractable complexity in modeling and theoretical analysis, i.e., each user is associated with at most one AP, either WiFi AP or LiFi AP. 
The user association vector $\textbf{u}_k(t)$ for user $k$ at time slot $t$ is denoted as $\textbf{u}_k(t) = \{u_{l,k}(t), u_{w,k}(t)|\forall{l}\in \mathcal{L}, \forall{w}\in \mathcal{W} \}$, where $u_{l,k}(t)$ and $u_{w,k}(t)$ are binary variables, with $u_{l,k}(t)=1$ or $u_{w,k}(t)=1$ 
when user $k$ is associated with LiFi AP $l$ or WiFi AP $w$, respectively, $0$ otherwise. 
Then, 
we can define the user-AP association vector for the entire hybrid network as $\textbf{u}(t) = \{\textbf{u}_k(t)| \forall{k}\in \mathcal{K}\}$. 

Once the connections between the users and APs are determined, the sum data rate of the hybrid networks at time slot $t$ can be written as:

\begin{align} \label{eq:sumdaterate}
r_{t} = &{\sum\limits_{l=1}^{\mathcal{L}}\sum\limits_{k=1}^{\mathcal{K}}u_{l,k}(t)r^{lf}_{l,k}(t)}+{\sum\limits_{w=1}^{\mathcal{W}}\sum\limits_{k=1}^{\mathcal{K}}u_{w,k}(t)r^{wf}_{w,k}(t)}. 
\end{align}

\subsection{Random Waypoint Model} \label{sec:sec:RWP}
In this work, we use the random waypoint model \cite{johnson1996dynamic} to model users' indoor moving trajectories, as it effectively replicates users' motion patterns across a range of indoor environments.
This model is extensively utilized in past research \cite{ahmad2020reinforcement, wu2019mobility}. Within the framework of the RWP model, individuals select a destination point at random within the designated area and proceed towards it in a direct line. The speed of each individual is also randomly determined, adhering to a uniform distribution within a specified range of minimum and maximum velocities. Upon reaching the destination, an individual may either pause for a set duration or resume movement, adhering to the previously described mechanism. This model accommodates the coexistence of mobile and stationary users within the environment at any given moment.

\section{Problem Formulation} \label{sec:problem}
The network control objective can be stated as maximizing the sum data rate of the hybrid indoor LiFi/WiFi downlink access networks by jointly considering the users' positions, 
user-AP association variables, as well as user moving patterns subject to bandwidth limitations of the hybrid networks. Define $\bm{\mathcal P}(t) = [\mathcal{P}_1(t), \mathcal{P}_2(t), \ldots, \mathcal{P}_K(t)]$ as the user location at time slot $t$, where height is assumed constant for all users and only $x$ coordinates and $y$ coordinates are considered to be moving over time, with $\mathcal{P}_k(t)=\{x_k(t), y_k(t)\}$. The network control problem can then be formulated as:

\noindent\text{\textbf{Problem:} }\text{Given:}\;\bm{\mathcal{P}}(t)
\begin{align} 
&\underset{\textbf{u}(t)}{\text{Maximize}} \quad f = r_{t} \label{eq:obj}\\
&\text{Subject to:}\;\;\; 0 \le x_k(t) \le X, \forall{k}\in \mathcal{K} \label{cs:locationx}, \\
&\qquad \qquad \quad \;0 \le y_k(t) \le Y, \forall{k}\in \mathcal{K} \label{cs:locationy},\\
&\qquad \qquad \quad \;{u}_{l,k}(t)\in \{0,1\}, \forall{k}\in \mathcal{K}, \forall{l}\in \mathcal{L} \label{cs:lfbinary},\\
&\qquad \qquad \quad \;{u}_{w,k}(t)\in \{0,1\}, \forall{k}\in \mathcal{K}, \forall{w}\in \mathcal{W}\label{cs:wfbinary},\\
&\qquad \qquad \quad \;\textbf{1}^{T}\textbf{u}_k(t)\le 1, \forall{k}\in \mathcal{K}\label{cs:nooverlap},\\
&\qquad \qquad \quad \;\underset{k\in K}{\sum}u_{l,k}(t) \le N_{l}, \forall{l}\in \mathcal{L} \label{cs:lf},\\
&\qquad \qquad \quad \;\underset{k\in K}{\sum}u_{w,k}(t) \le N_{w}, \forall{w}\in \mathcal{W}. \label{cs:wf}
\end{align}

In the above-formulated problem, constraints (\ref{cs:locationx}) and (\ref{cs:locationy}) restrict that all the users should be within the predefined indoor area.  
Constraints (\ref{cs:lfbinary}) and (\ref{cs:wfbinary}) require that the user-AP association variables are binary.
(\ref{cs:nooverlap}) indicates that each user can be only served by one AP. 
The capacity limitations (aka the maximum number of connected users) of LiFi and WiFi APs are set in (\ref{cs:lf}) and (\ref{cs:wf}). We consider maximizing the sum throughput in the objective function, however, without loss of generality, our method can be extended to incorporate additional factors such as energy consumption, latency, or fairness.
\begin{corollary} Problem 1 is a non-convex binary integer programming (BIP) problem. \end{corollary} 
\begin{IEEEproof}From the constraints (\ref{cs:lfbinary}) and (\ref{cs:wfbinary}), the decision variables $u_{l,k}(t)$ and $u_{w,k}(t)$ are binary. Moreover, $r_{t}$ in the objective function (\ref{eq:obj}) is the summation of individual throughput terms $r_{l,k}^{lf}$ and $r_{w,k}^{wf}$ in (\ref{eq:lfdatarate}) and (\ref{eq:wfdatarate}), which are independent of variables $u_{l,k}(t)$ and $u_{w,k}(t)$. Thus, the objective function $r_t$ is a weighted linear summation of binary variables. Additionally, the single-home assignment constraint (\ref{cs:nooverlap}) and the capacity limitation constraints (\ref{cs:lf}) and (\ref{cs:wf}) are linear. Therefore, Problem 1 is classified as a binary integer programming problem \cite{boyd2004convex}. Moreover, the feasible set of Problem 1 is a non-convex set due to binary constraints, thereby we conclude this BIP problem is non-convex. \end{IEEEproof}

\begin{figure*}[t]
\begin{center}
\includegraphics[width=0.9\textwidth]{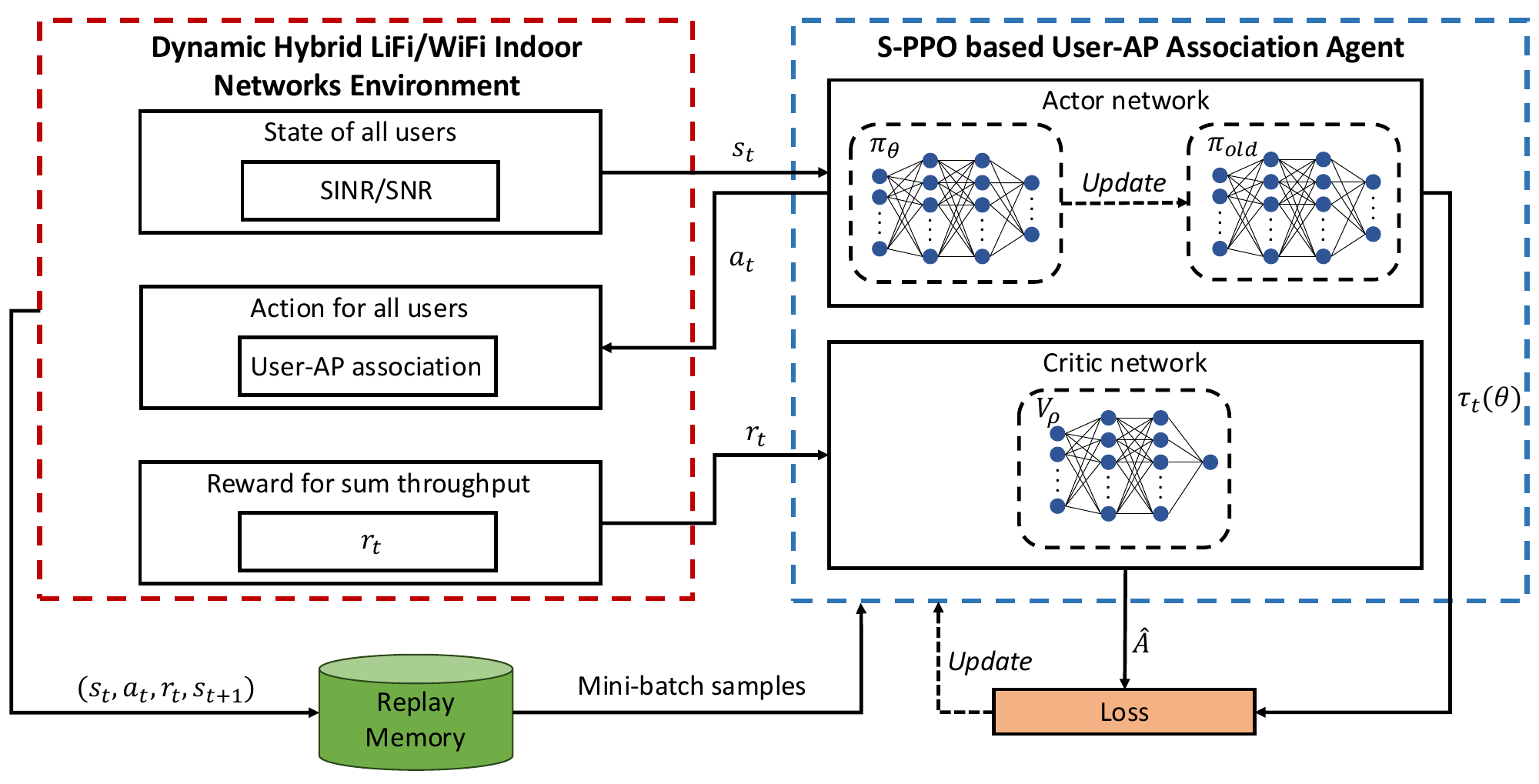}
\caption{\small{Actor-Critic S-PPO based user association scheme.}} 
\vspace{-6mm}
\label{fig:PPO}
\end{center}
\end{figure*}

\section{Proposed S-PPO Method} \label{sec:solu}
Problem 1 is a non-convex BIP problem, for which there are generally no existing solution algorithms that can obtain the global optimum in polynomial computational time. 
Traditional solution algorithms face significant challenges in solving this problem because they are computationally intensive due to the dynamic features of the wireless environment and therefore cannot effectively obtain the global optimal solution. In contrast, learning-based methods can reformulate the optimization challenge into a decision-making problem, showing greater flexibility in dynamic environments. Therefore, in this paper, we propose a sequential-proximal policy optimization method as a solution algorithm for the formulated NP-hard problem, where an action space decomposition method is designed to further significantly reduce computational complexity.

\subsection{Preliminaries} \label{sec:sec:pre}

In this section, we introduce the preliminaries of the proximal policy optimization algorithm. The PPO algorithm is based on the actor-critic framework, which integrates the strengths of both value-based (such as Q-learning) and policy-based (such as gradient descent) approaches in reinforcement learning. This integration
enables quicker convergence and more stable performance. Additionally, PPO is less sensitive to the changes in hyperparameters, which makes the tuning process more efficient, thus allowing our proposed PPO-based solution method can be seamlessly applied to various network topologies.

In the proposed S-PPO, the actor-critic framework has two networks, as shown in Fig. \ref{fig:PPO}, the policy network $\pi_{\theta_t}$ (actor) and the value network $V_{\rho_t}$ (critic), where $\theta_t$ and $\rho_t$ are the parameters of these two networks at time slot $t$ \cite{konda1999actor}. Given a state $s_t$, the probability of the policy $\pi_{\theta_t}$ executes action $a_t$ denoted as $\pi_{\theta_t}(a_t|s_t)$. The underlying principle of the PPO algorithm is to update the policy in a manner that maximizes expected reward without drastic policy changes, which is achieved by the clipped surrogate objective function \cite{schulman2017proximal}:

\begin{align} \label{eq:loss}
L^{CLIP}(\theta)=\hat{\mathbb{E}}_t[min(\tau_t({\theta})\hat{A_t},clip(\tau_t(\theta),1-\epsilon,1+\epsilon)\hat{A_t})], 
\end{align}where $\tau_t(\theta)$ is the probability ratio between the new and old policies. This ratio measures how the new policy deviates from the old one, which can be calculated as:

\begin{align} \label{eq:prob}
\tau_t(\theta)=\frac{\pi_\theta(a_t|s_t)}{\pi_{\theta_{old}}(a_t|s_t)}. 
\end{align}Moreover, $\hat{A_t}$, 
the generalized advantage estimator, is used to measure how much better it is to take a specific action compared to the average action at a given state, 
as:

\begin{align} \label{eq:adv}
\hat{A}_t=R_t-V_{\rho}(s_t),
\end{align}
with $R_t$ representing the accumulative reward, $V_{\rho}(s_t)$ denoting the expected return of the state $s_t$. 

Additionally, the clip function is defined as:

\begin{align} \label{eq:clip}
clip(\tau_t(\theta),1-\epsilon,1+\epsilon) = \begin{cases}
       1-\epsilon \; & \tau_t(\theta) < 1-\epsilon,  \\
       1+\epsilon \; & \tau_t(\theta) > 1-\epsilon,  \\
       \tau_t(\theta) \; & \text{otherwise},  
      \end{cases}
\end{align}where $\epsilon$ is a clipping hyperparameter.

On the other hand, the loss function to update the critic network is typically a mean squared error (MSE) loss, which can be given as:

\begin{align} \label{eq:critic}
L^{Critic}(\rho) = \frac{1}{N} \sum_{i=1}^{N}(V_{\rho}(s_i) - R_i)^2,
\end{align}where $N$ is the number of state-value pairs in the batch used for the update.

In summary, at state $s_t$, the actor network is used to select actions $a_t$ based on the current policy $\pi_\theta$ to interact with the environment, and then receive reward $r_{t}$. Next, the experience tuples $\{s_t, a_t, r_{t}, s_{t+1}\}$ will be stored in the memory. After several timesteps, the data in the memory will be used to calculate the loss. The actor and critic network will then be updated by the gradient of $L^{CLIP}(\theta)$ and $L^{Critic}(\rho)$, respectively, thus obtaining the optimal policy.

\subsection{Sequential-Proximal Policy Optimization} \label{sec:sec:sppo}

Next, we present the framework of the proposed RL method and action space decomposition method in detail as follows.

\noindent\textbf{Agent}: The agent is a central unit that can obtain the user positions, and execute the proposed network control algorithm to accomplish the user-AP association. \\
\noindent\textbf{Sequential actions}: In RL, the high dimensionality of the action space will have a negative impact on sample efficiency and model expressiveness \cite{10228875}. The potential action space, denoted as $K^{(L+W)}$, expands exponentially with the growth of user number $K$, and the numbers of APs $L$ and $W$, respectively. To address the issue of this exponentially increasing action space, 
we design a new action space decomposition approach. 
The key idea is to divide the action space into $K$ sequential actions for each user. This approach effectively breaks down the overall association policy into a sequencing policy to determine the action for each user sequentially. After all user-AP association actions are collected, they are composed into a combined action and then sent to the wireless network environment. In this way, the overall action space is reduced to $K(L+W)$.

The action taken at time slot $t$ for each user is determined by the current policy network, expressed as:

\begin{align} \label{eq:action1}
    a_t^k= [\textbf{u}_{k}(t)], a_t^k \in \mathcal{A}, k\in\mathcal{K},
    \end{align}which is the user-AP association vector of the hybrid LiFi/WiFi networks for user $k$. $\mathcal{A}$ is the action space including all possible actions. Then the combined action for all the users is defined as:

\begin{align} \label{eq:action2}
    \textbf{a}_t= [\textbf{u}(t)].
    \end{align} 
    
\noindent\textbf{States}: To obtain the individual user action, the state is defined as the received SINR/SNR from LiFi/WiFi APs of each user:

\begin{align} \label{eq:state}
    s^k_t= [\bm{\gamma}^k(t)], s^k_t \in \mathcal{S},  k\in\mathcal{K},
    \end{align}where $\bm{\gamma}^k(t) = \{\gamma_{l,k}^{lf}(t), \gamma_{w,k}^{wf}(t)\}, l\in\mathcal{L}, w\in\mathcal{W}$, and 
    $\mathcal{S}$ denotes the state space. The combined state for all users can be given by:
\begin{align} \label{eq:state2}
    \bm{s}_t= [\bm{\gamma}(t)].
    \end{align}

\noindent\textbf{Policy}:
The stochastic policy $\pi_{\theta}(a^k_t|s^k_t)$ is the probability of choosing action $a^k_t$ at state $s^k_t$, which can be defined as:

    \begin{align} \label{eq:policy}
    \pi_{\theta}(a^k_{t}|s^k_{t})=\Pr(a^k_t|s^k_t).
    \end{align}By selecting the actions according to the current policy and interacting with the environment over time, we then can obtain the state–action sequence trajectory as $\{s_0,a_0,s_1,a_1,\cdot\cdot\cdot,s_T\}$. 
    
\noindent\textbf{Reward}: At time slot $t$, when action $\bm{a}_t$ is chosen at state $\bm{s}_t$, the immediate reward received is the sum data rate of the networks $r_{t}$. Therefore, the discounted accumulative reward of a trajectory can be expressed as:

    \begin{align} \label{eq:reward}
    R_t=\sum\limits_{t=0}^{T}\beta^{t}r_{t},
    \end{align}where $\beta\in(0,1)$ is the discount factor, which determines the effect of the current immediate reward on the future sum rewards. To ensure that the trained solutions satisfy AP capacity constraints (\ref{cs:lf}) and (\ref{cs:wf}), we assign a larger penalty when the number of users connected to AP exceeds the limitation, thus forcing the agent to avoid infeasible solutions. 
The detailed S-PPO method is shown in Algorithm \ref{alg:sppo}.


\RestyleAlgo{ruled}

\begin{algorithm}[t]
  \caption{S-PPO for user association problem in hybrid LiFi/WiFi indoor networks}\label{alg:sppo}

  \KwIn{Initialize the parameters $\theta_0$ of policy $\pi$ (i.e., actor network), the parameters $\rho_0$ of the value function $V$ (i.e., critic network).}
  \KwOut{Optimal policy $\pi^*$}

  \While{episode $<$ max episode}{
    \While{timestep $<$ update interval}{
        \For{each user}{
            Select action $a_k(t)$ according to policy $\pi_{\theta_t}$\;
        }
        Environment calculates reward $R_t$ based on combined action $\textbf{a}_t$\;
        Observe new combined state $\bm{s}_t$ based on local observations and previous experience\;
        Store experience tuple into memory\;
    }
    Update policy $\pi$ with the clipped surrogate objective (\ref{eq:loss})\;
    Update value function $V$ with the loss function (\ref{eq:critic})\;
    Clear memory\;
    timestep $\gets$ 0\; 
    }
\end{algorithm}
 \vspace{-2mm}

\section{Performance Evaluation} \label{sec:performance}

In this section, we evaluate the effectiveness of the proposed S-PPO based solution algorithm through comprehensive simulations. 

\subsection{Simulation Setup} \label{sec:sec:simulation}

\begin{figure}[t]
\centering 
\vspace{-4.5mm}
\subfigure[Interference-free scenario]{
\label{fig:scenario1}
\includegraphics[width=4.3cm,]{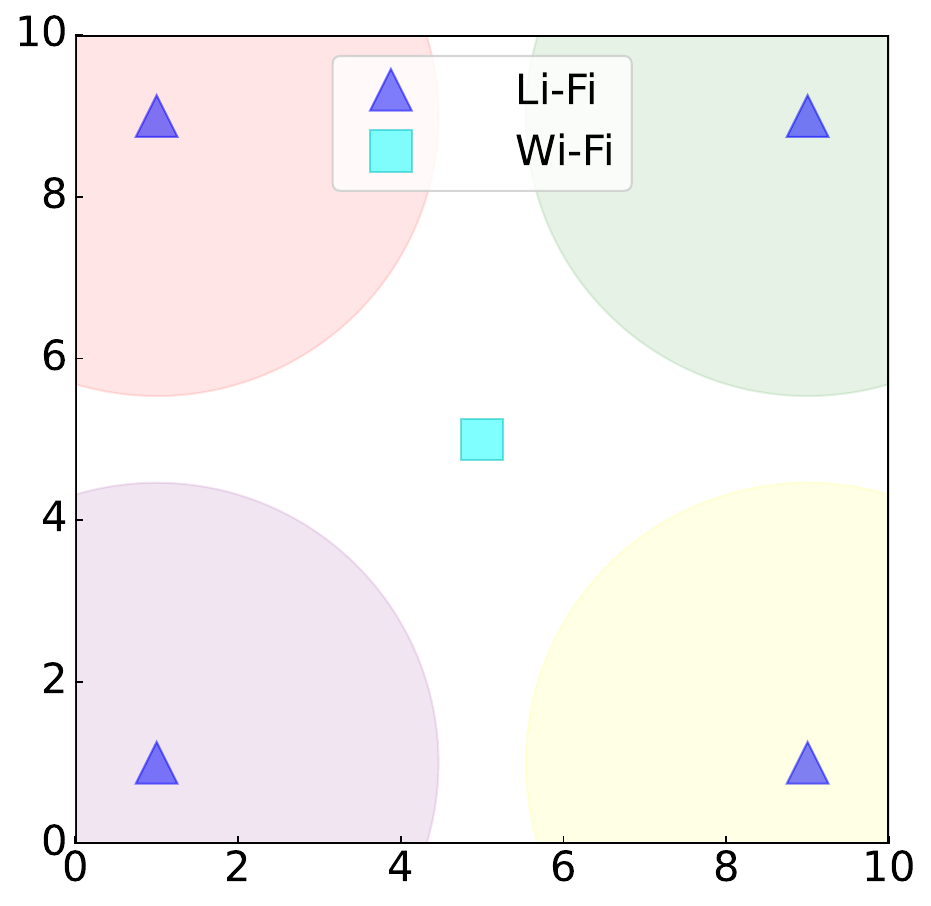}}\subfigure[Interference-prone scenario]{
\label{fig:scenario2}
\includegraphics[width=4.3cm,]{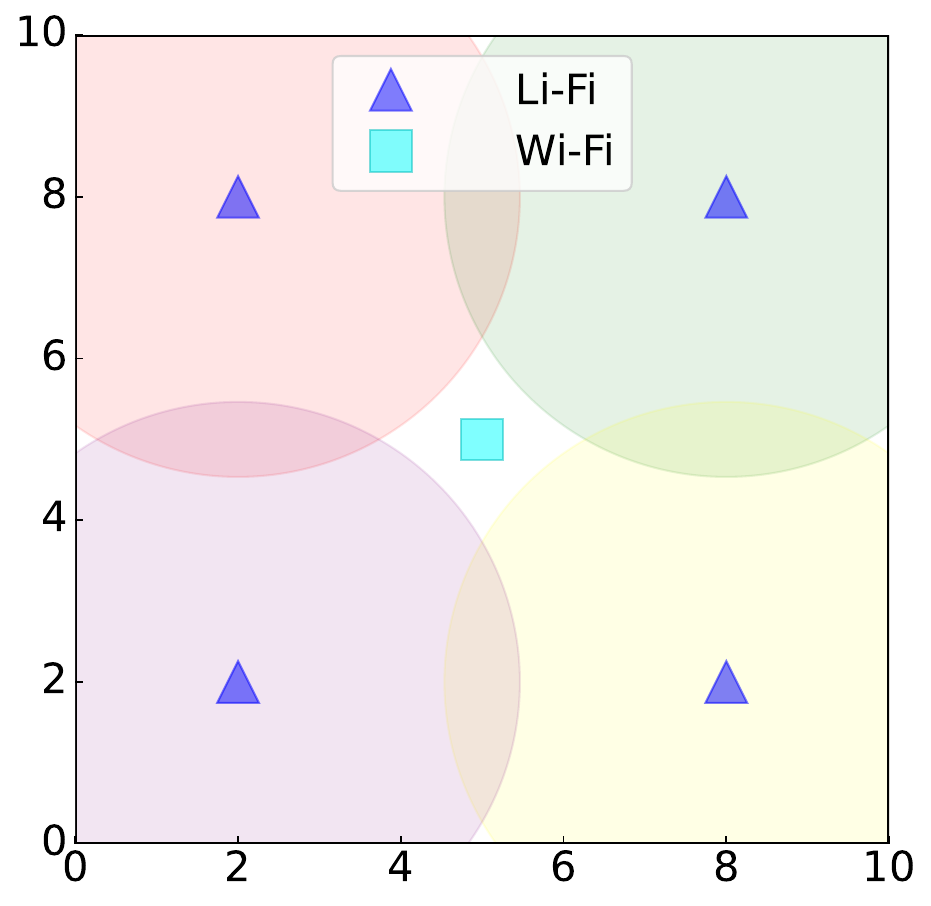}}
\caption{\small{Hybrid LiFi/WiFi networks topology.}}
\vspace{-4mm}
\label{fig:scenario}
\end{figure}

We consider an indoor room area of $10 \times 10 \times 3.5$ $\text{m}^3$, as illustrated in Fig. \ref{fig:real}. To study 
the impact of inter-cell interference from LiFi APs on the sum throughput of the hybrid network, two representative AP configurations are evaluated: 
interference-free scenario and interference-prone scenario, where the numbers of AP and user range from 5 to 7 and 2 to 18, respectively. Example topologies are depicted in Fig. \ref{fig:scenario}. Figure \ref{fig:scenario1} shows an interference-free scenario configuration, which employs a layout of $L = 4$ LiFi APs located at the four corners of the room, and $W = 1$ WiFi AP 
at the center. The circled areas with different colors indicate the coverage areas of different VLC APs, while the WiFi AP covers the entire room. In contrast, the interference-prone scenario as presented in Fig. \ref{fig:scenario2}, has the same number of APs as the first scenario. However, in this scenario, the LiFi APs are positioned closer to each other to enhance coverage within the room. Consequently, it leads to overlapping coverage areas, increasing the likelihood of inter-cell LiFi interference. Without loss of generality, our proposed S-PPO method can be applied to other scenarios with different numbers of APs and users.
The parameters of LiFi and WiFi networks are summarized in Table~\ref{tb:lifi} and Table~\ref{tb:wifi}, respectively. 

To validate the proposed method, two different network settings with different network capacity constraints are considered: (i) networking setting 1: $N_l = 2, N_w = 5$;
(ii) networking setting 2: $N_l = 3, N_w = 7$. 
The user location and height are generated randomly. For each networking setting, static user scenario and mobile user scenario are studied, respectively.
In the mobile user scenario, as discussed before, the RWP model is used, where the user moving speed is randomly selected between $0.5$ m/s and $2$ m/s. The user location information is obtained every $0.1$ s, ensuring that the users' locations are tracked in a nearly real-time manner yet do not change significantly. 
The RL parameters used in the simulation are summarized in Table \ref{tb:rl}.

\small
\begin{table}[b]
\scriptsize{
\vspace{-4mm}
\caption{LiFi Channel Parameters} \label{tb:lifi}
\begin{center}
\begin{tabular}{ |c|c|c| } 
 \hline
 Parameter & Symbol & Value \\
 \hline
 Height between ceiling and user & $h$ & 1.5 m - 2 m \\ 
 \hline
 The physical area of a PD & $A$ & $1 \, \text{cm}^2$ \\
 \hline
 The gain of the optical filter & $T_s(\psi_{l,k})$ & 1.0 \\ 
 \hline
 Receiver FoV semi-angle & $\Psi_{1/2}$ & 90 degree \\ 
 \hline
 Refractive Index & $n$ & 1.5 \\ 
 \hline
 Half-intensity radiation angle & $\Phi_{1/2}$ & 60 degree \\
 \hline
 Detector responsivity & $\kappa$ & 0.53 A/W \\
 \hline
 Transmit optical power per LiFi AP & $P_{lf}$ & 3 Watt \\
 \hline
 Noise power spectral density & $\mathcal{N}_{lf}$ & $10^{-21} \, \text{A}^2/\text{Hz}$ \\
 \hline
 Bandwidth per LiFi AP & $B_{lf}$ & 40 MHz \\
 \hline
 
\end{tabular}
\end{center}}
\end{table}
\normalsize

\small
\begin{table}[b]
\scriptsize{
\vspace{-4mm}
\caption{WiFi Channel Parameters} \label{tb:wifi}
\begin{center}
\begin{tabular}{ |c|c|c| } 
 \hline
 Parameter & Symbol & Value \\
 \hline
 Breakpoint distance & $d_{BP}$ & 5 m\\ 
 \hline
 Central carrier frequency & $f_c$ & 2.4 GHz \\ 

 \hline
 Transmit power & $P_{wf}$ & 0.1 Watt \\ 
 \hline
 Bandwidth per WiFi channel & $B_{wf}$ & 10 MHz \\ 
 \hline
 Noise power spectral density & $\mathcal{N}_{wf}$ & $4.002 \times 10^{-17} \, \text{W/Hz}$ \\

 \hline
\end{tabular}
\end{center}}
\end{table}
\normalsize

\small
\begin{table}[b]
\scriptsize{
\vspace{-4mm}
\caption{S-PPO hyperparameters} \label{tb:rl}
\begin{center}
\begin{tabular}{ |c|c| } 
 \hline
 Parameter & Value \\
 \hline
 Learning rate & 0.0001 \\ 
 \hline
 Clipping coefficient ($\epsilon$) & 0.1 \\ 
 \hline
 Discount factor ($\beta$) & 0.99 \\ 
 \hline
 Number of epoch per iteration & 8 \\ 
 \hline
 Update interval & 450 \\
 \hline
\end{tabular}
\vspace{-5mm}
\end{center}}
\end{table}
\normalsize

\subsection{Compared Methods} \label{sec:sec:compare}
\noindent\textbf{Exhaustive search}: Exhaustive search is a brute-force approach to solve optimization problems by evaluating all possible solutions. For the proposed problem, ES is used to enumerate all possible AP-user association cases to find the globally optimal AP-user associations, which will provide the upper bound of our work but at the cost of significant computational complexity. The objective function of exhaustive search is the same as (\ref{eq:obj}).

\noindent\textbf{Signal strength strategy}: SSS is a widely adopted technique in wireless communication, focusing on selecting the AP that provides the highest signal-to-noise ratio for the users \cite{wu2017access, wu2019mobility}. When user $k$ is served by AP $i$ at time slot $t$, the received SNR $\sigma_{i,k}$ is denoted as:

\begin{align} \label{eq:snr}
\sigma_{i,k}(t) &=\begin{cases}
      \frac{(\kappa H^{lf}_{l,k}(t)P_{lf})^2}{\mathcal{N}_{lf}B_{lf}}\; & \text{if}\ i\ \text{is LiFi AP},  \\ 
      \frac{|H_{w,k}^{wf}(t)|^2P_{wf}}{\mathcal{N}_{wf}B_{wf}} \; & \text{if}\ i\ \text{is WiFi AP}.
      \end{cases}\; 
\end{align}Then, the objective function of SSS method for user $k$ can be written as:

\begin{equation}\begin{aligned} \label{eq:sss}
\underset{\textbf{u}_k(t)}\max\ &\sigma_{i,k}(t) \\
\text{s.t.}\ & i \in{\mathcal{L}\cup \mathcal{W}},\; \\
& k \in{\mathcal{K}}.
\end{aligned}\end{equation}

\noindent\textbf{Trust Region Policy Optimization}: TRPO is chosen for comparison due to its application in addressing the user-AP association problem within hybrid LiFi/WiFi networks \cite{ahmad2020reinforcement}. It operates within the same structural framework as S-PPO, utilizing actor and critic networks. However, TRPO differentiates itself by implementing policy updates through trust region constraints, while PPO ensures stability using a clipped
function. 
The objective function of TRPO is the same as (\ref{eq:obj}). The learning rate and discount factor for TRPO are the same as those for S-PPO. 

\subsection{Complexity Analysis} \label{sec:sec:complex}
As discussed in Sec. \ref{sec:solu}, the proposed S-PPO framework consists of actor and critic networks, with each being a 2-hidden-layer deep neural network, where each layer has $N_h$ neurons. The computational complexity of the actor network to process one user is $\mathcal{O}(2(L+W)N_h+N_h^2)$. Then for the $K$ user, it is $\mathcal{O}(K(2(L+W)N_h+N_h^2))$. While the critic network has a complexity of $\mathcal{O}((L+W)N_h+N_h^2+N_h)$. Consequently, the aggregate computational complexity of the S-PPO methodology is calculated as $\mathcal{O}((2K+1)(L+W)N_h+(K+1)N_h^2+N_h)$.
Due to identical network architectures, the TRPO method shares the same complexity level as S-PPO. In comparison, the computational complexities for ES and SSS techniques for user-AP association are $\mathcal{O}((L+W)^K)$ and $\mathcal{O}((L+W)K)$, respectively.


\begin{figure}[t]
\centering 
\subfigure[Network setting 1]{
\label{fig:static1}
\includegraphics[width=4.3cm,height = 3cm
]{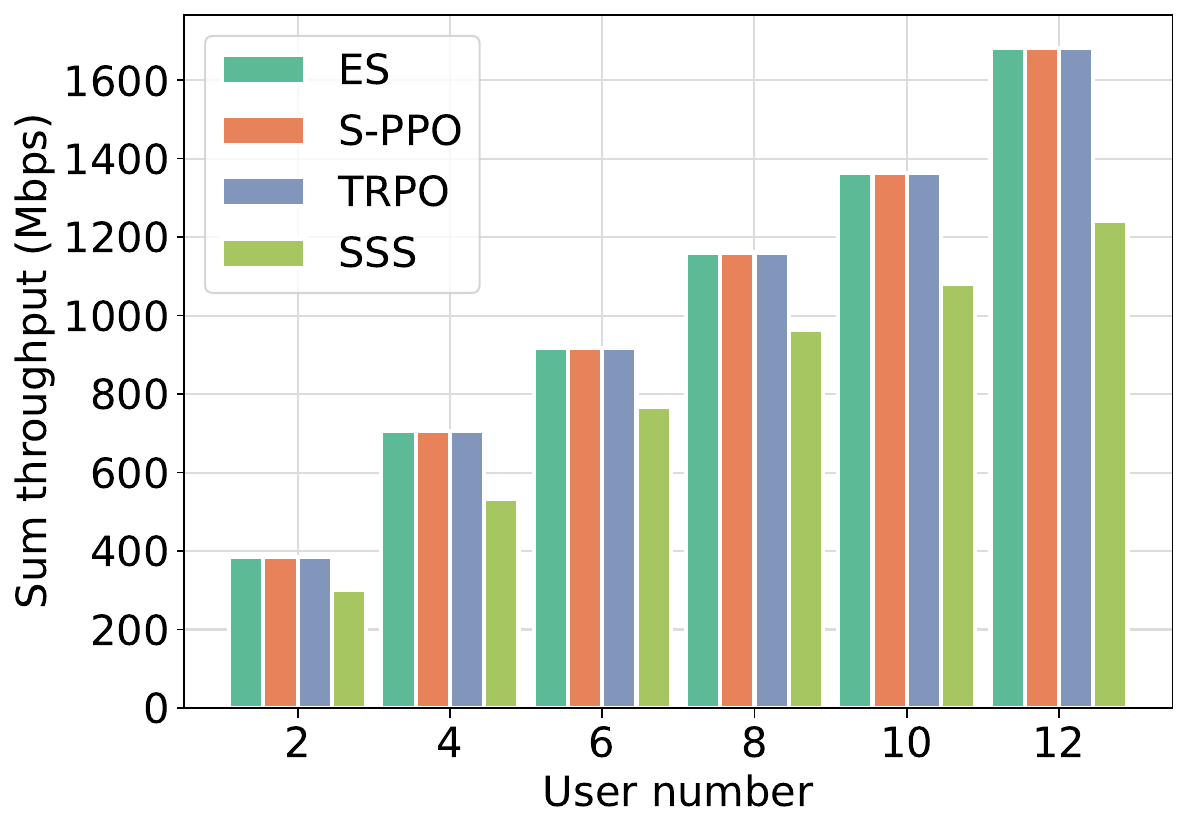}}\subfigure[Network setting 2]{
\label{fig:static2}
\includegraphics[width=4.3cm,height = 3cm
]{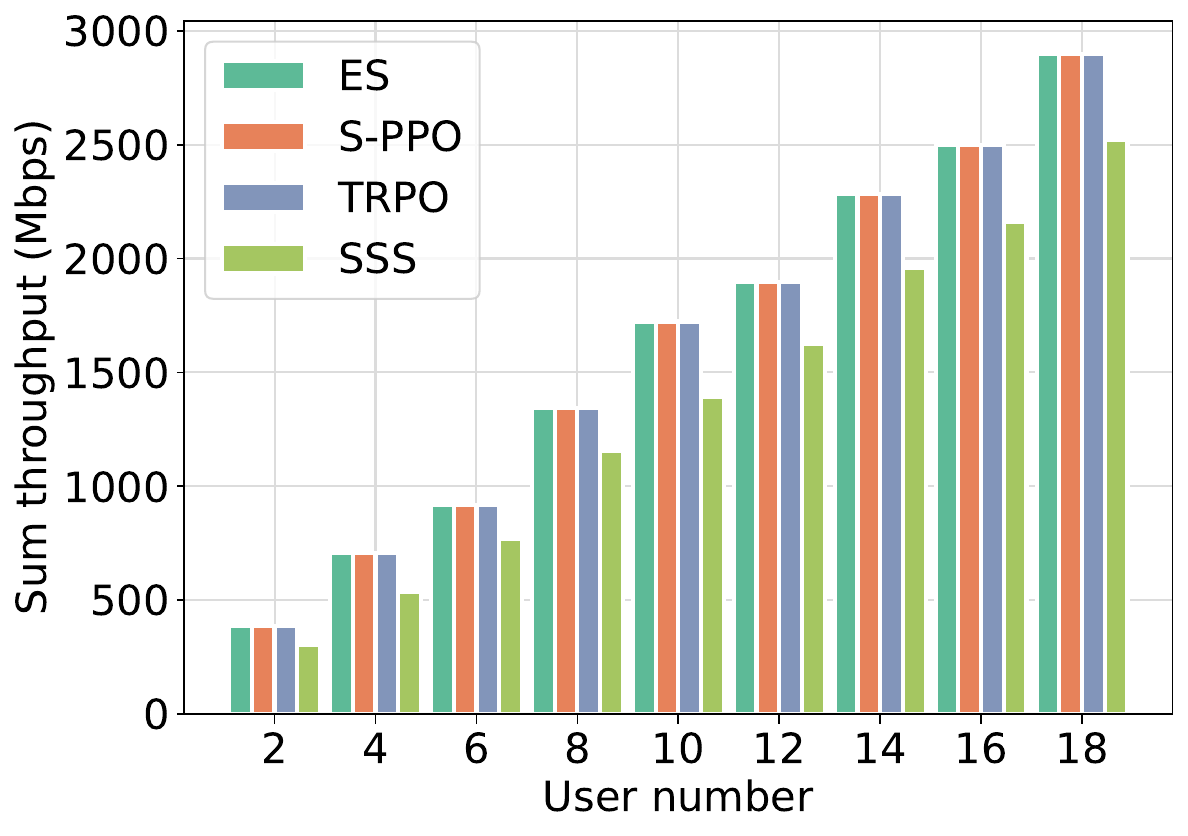}}
\caption{\small {Sum data rate versus user number in static interference-prone scenario.}}
\vspace{-4mm}
\label{fig:static}
\end{figure}

\begin{figure}[t]
\centering 
\subfigure[Static scenario]{
\label{fig:convstatic_free}
\includegraphics[width=4.3cm,height = 3cm
]{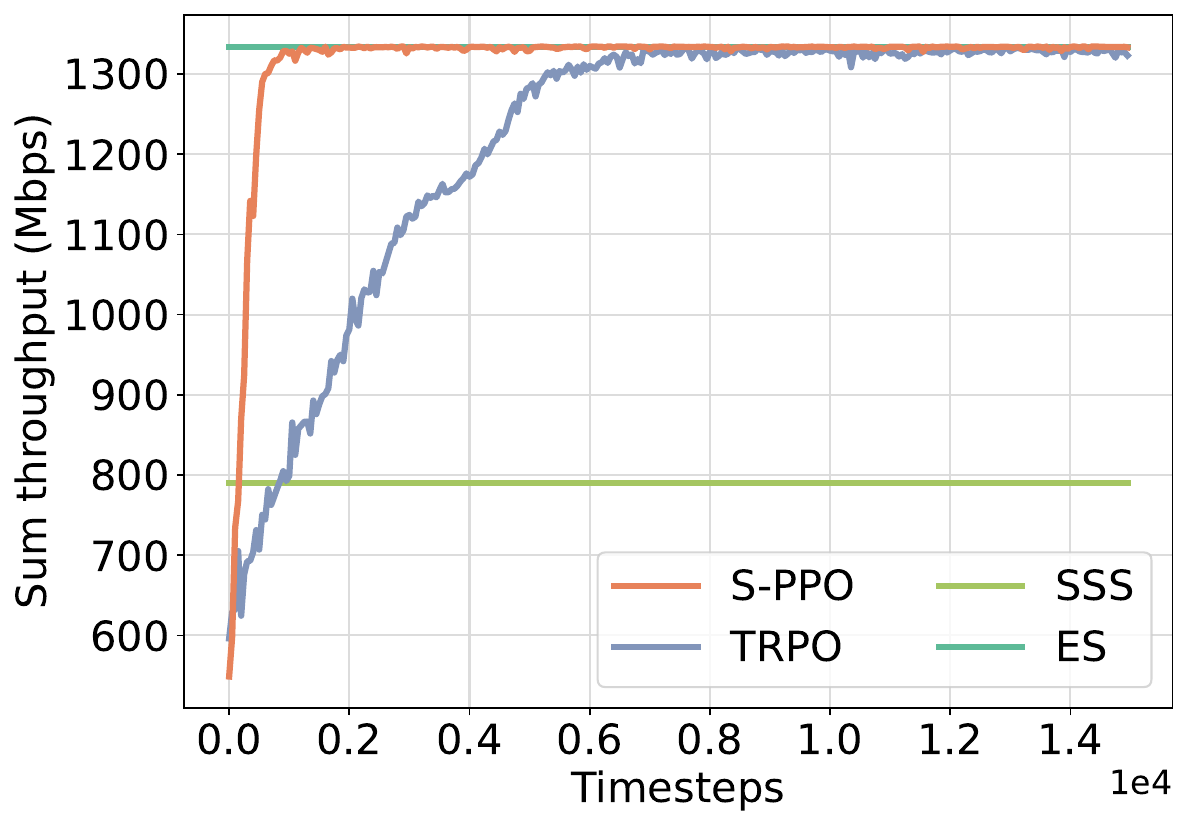}}\subfigure[Mobile scenario]{
\label{fig:convmobile_free}
\includegraphics[width=4.3cm,height = 3cm
]{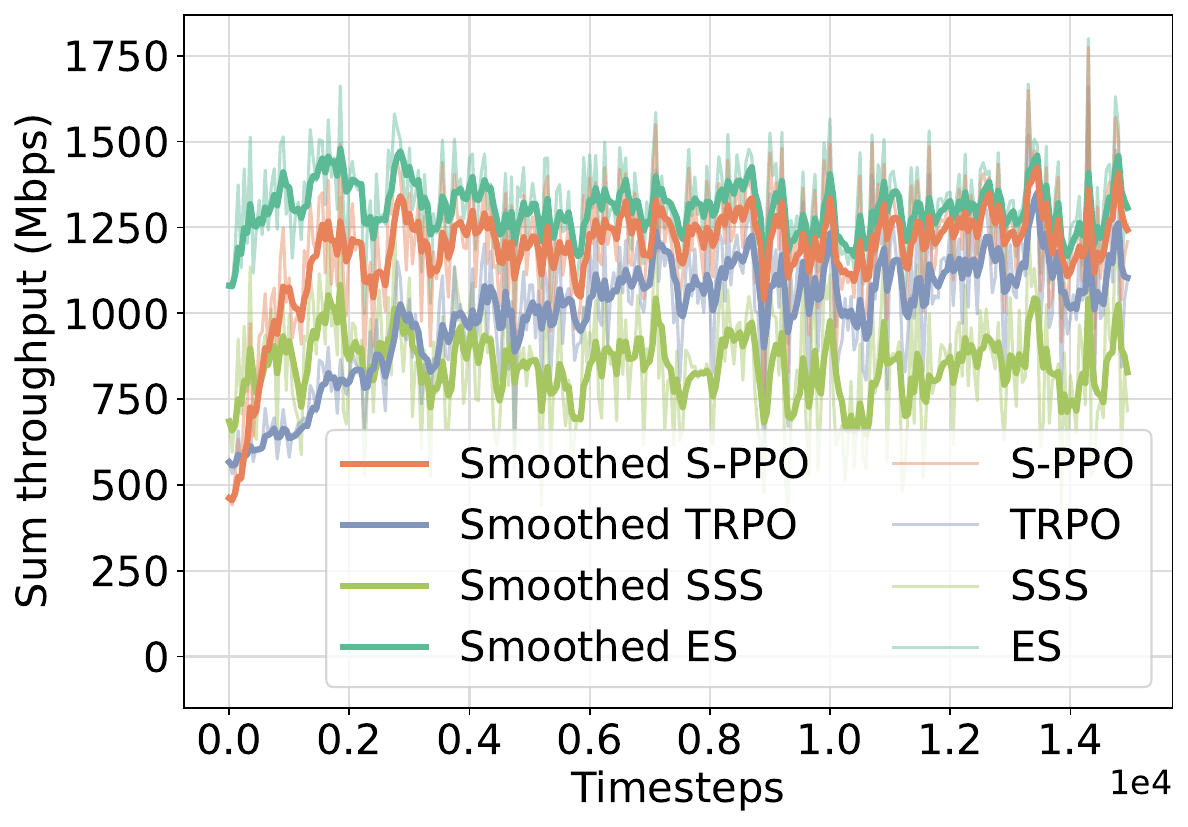}}
\caption{\small{Convergence comparison for 12 users in interference-free scenario, network setting 2.}}
\vspace{-6mm}
\label{fig:covergence_free}
\end{figure}

\begin{figure}[t]
\centering 
\subfigure[Static scenario]{
\label{fig:convstatic}
\includegraphics[width=4.3cm,height = 3cm
]{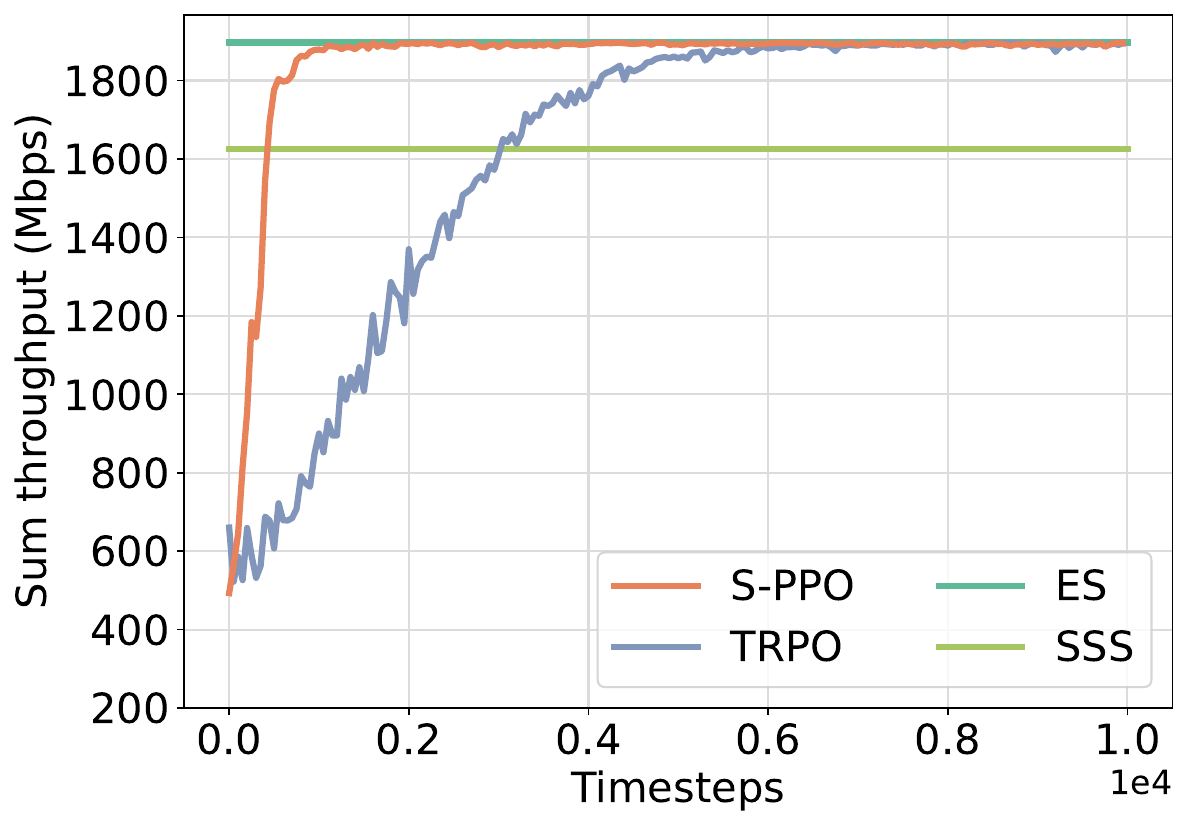}}\subfigure[Mobile scenario]{
\label{fig:convmobile}
\includegraphics[width=4.3cm,height = 3cm
]{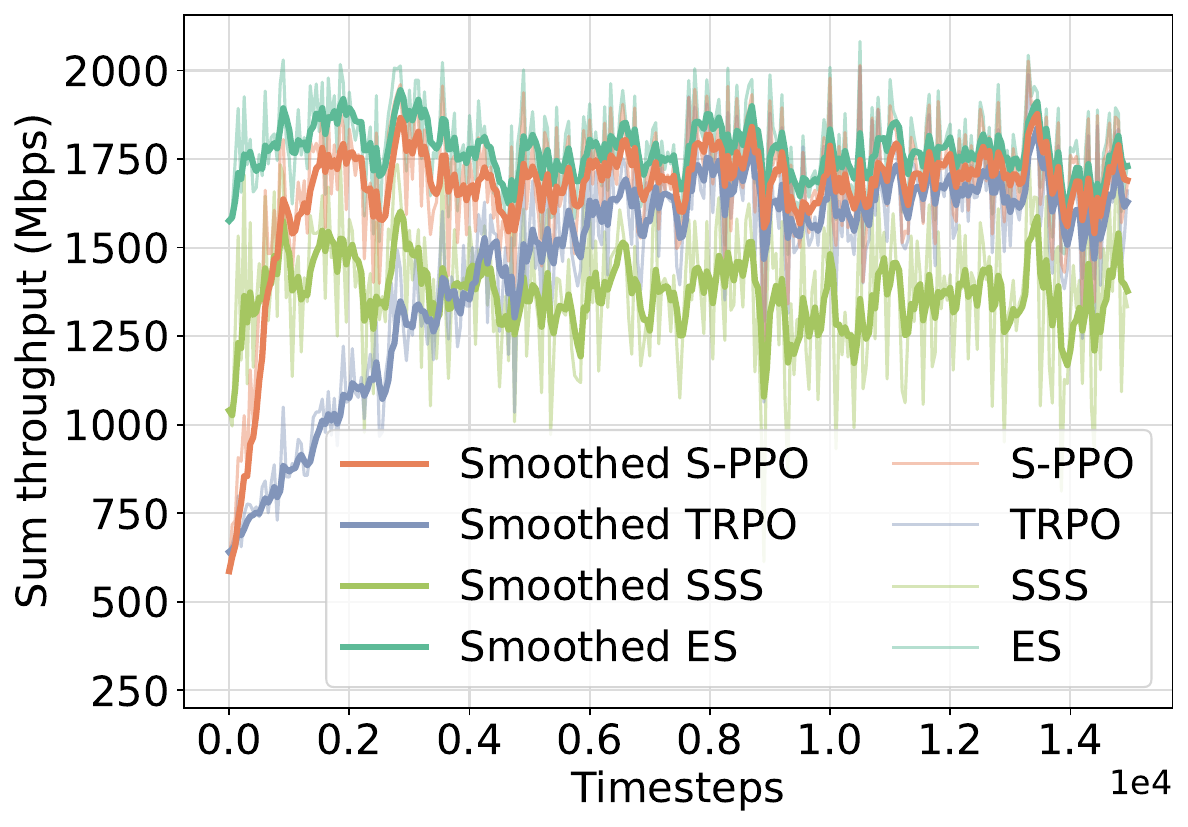}}
\caption{\small{Convergence comparison for 12 users in interference-prone scenario, network setting 2.}}
\vspace{-4mm}
\label{fig:covergence}
\end{figure}

\begin{figure}[t]
\centering 
\subfigure[Learning rate]{
\label{fig:robust_1}
\includegraphics[width=4.3cm,height = 3cm
]{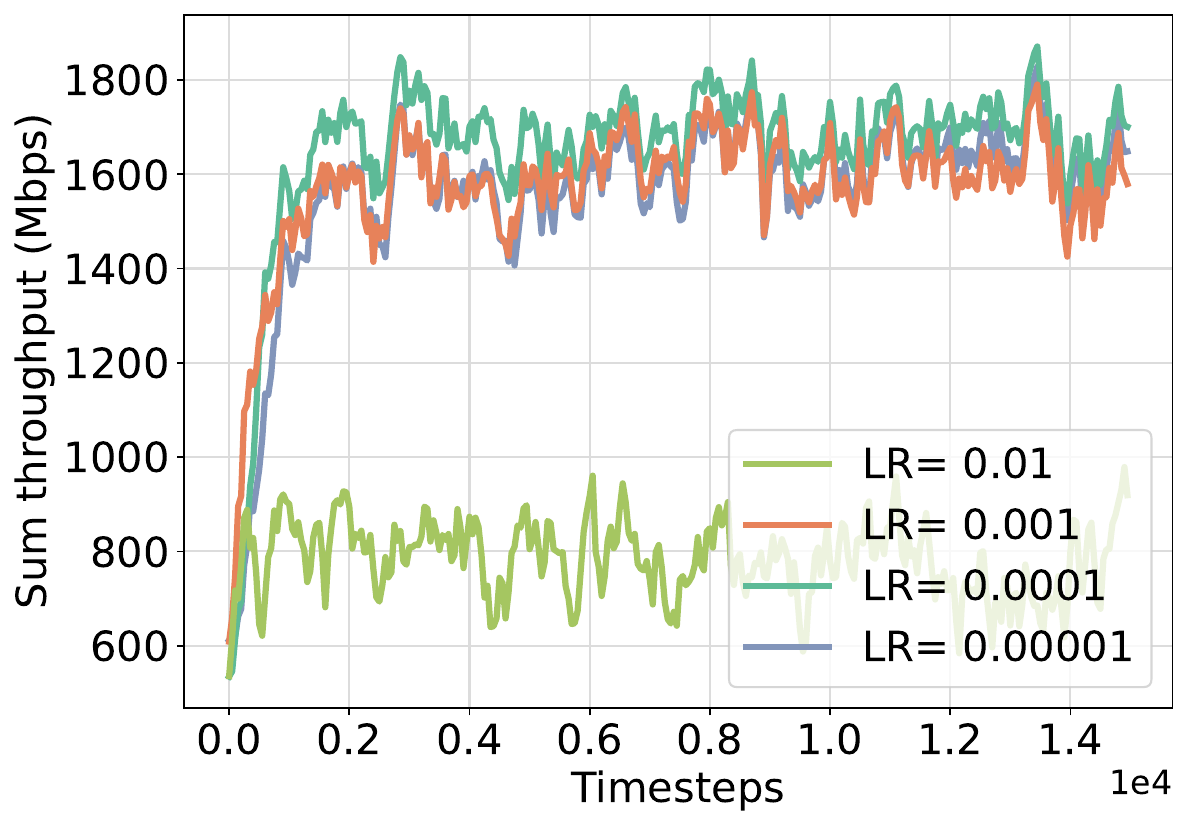}}\subfigure[Clip parameter]{
\label{fig:robust_2}
\includegraphics[width=4.3cm,height = 3cm
]{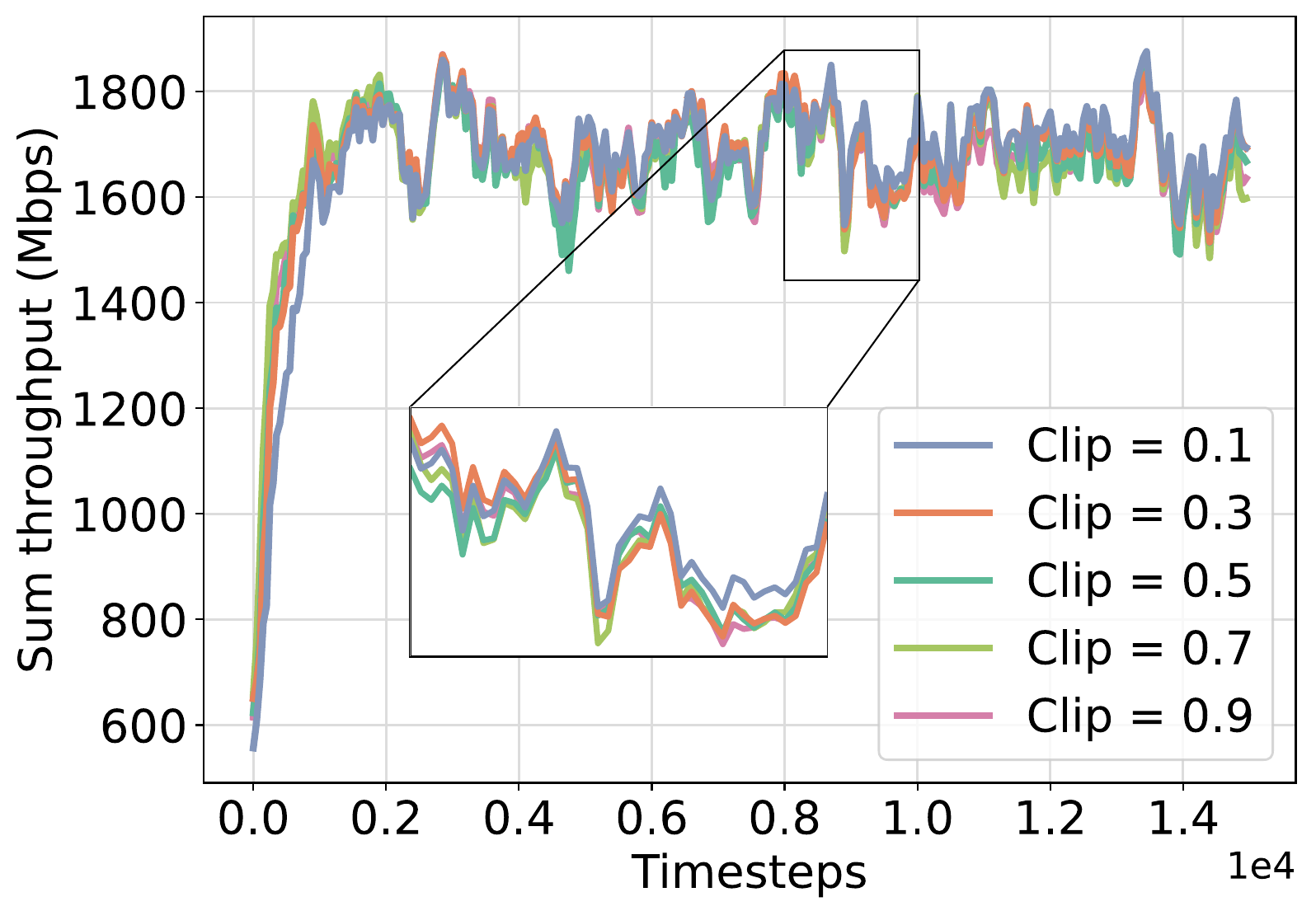}}
\caption{\small{Robustness comparison for 12 users in interference-prone scenario, network setting 2.}}
\label{fig:robust}
\end{figure}

\subsection{Simulation Results} \label{sec:testrslts} 
In this section, the performance of the proposed S-PPO method is evaluated by comparing it with the results obtained from ES, SSS, and TRPO in terms of optimality, convergence, robustness, computational complexity, scalability, and fairness under different network scenarios and settings. Additionally, we evaluate the fairness of the network.

\textbf{Optimality}: We first examine the optimality of the proposed S-PPO in static interference-prone networking scenarios with different capacity constraints by comparing the sum data rate with the other three methods. 
As illustrated in Fig. \ref{fig:static}, it is obvious that both the S-PPO method and TRPO can achieve the global optimal solution as the results achieved by the benchmark method ES. Moreover, S-PPO outperforms SSS by up to 35.5\% in both network settings. It can also be observed that as the user number increases, the performance of the proposed PPO will not degrade, thus validating its reliability.

\textbf{Convergence}: We then evaluate the convergence speed of the proposed S-PPO method under network setting 2 with 12 users, considering both interference-free and interference-prone scenarios in static and mobile cases.
In the static case, as Figs. \ref{fig:convstatic_free} and \ref{fig:convstatic} demonstrate, S-PPO converges 3 and 6 times faster than TRPO in interference-free and interference-prone scenarios, respectively. Moreover, both S-PPO and TRPO methods can achieve global optimal and outperform SSS by 16.7\% with respect to sum throughput in the interference-prone scenario. 

In the mobile scenario, to better illustrate the convergence trend, an exponential moving average is applied to the raw data. As depicted in Figs. \ref{fig:convmobile_free} and \ref{fig:convmobile}, the proposed S-PPO method demonstrates significantly faster convergence, achieving up to a 300\% improvement over TRPO. Moreover, S-PPO achieves near-optimal performance and outperforms both TRPO and SSS.

\textbf{Robustness}: 
To evaluate the robustness of the proposed S-PPO, we first conduct simulations to examine its sensitivity to changes in two important hyperparameters, including the learning rate (LR) and clipping parameter. Figure \ref{fig:robust_1} shows the impact of different learning rates on the performance of the proposed S-PPO. We can clearly see that the performance of S-PPO is not significantly affected in the particular range of LR from 0.00001 to 0.001. Though the performance is greatly degraded with a LR of 0.01, since it is not widely adopted, without loss of generality, we don't consider it a big drawback of the proposed S-PPO. The results of clipping parameters are shown in Fig. \ref{fig:robust_2}, where the clipping parameter ranges from 0.1 to 0.9 with an incremental interval of 0.2. The results remain consistent within the range, except for a light improvement observed at a clipping parameter of 0.1 compared with other values. Based on the results discussed above, we can conclude that the proposed S-PPO method shows its robustness against changes in learning rate and clipping parameter. 

\begin{figure}[t]
\centering 
\subfigure[Network setting 1 with 20\% blockage]{
\label{fig:block_1}
\includegraphics[width=4.3cm,height = 3cm
]{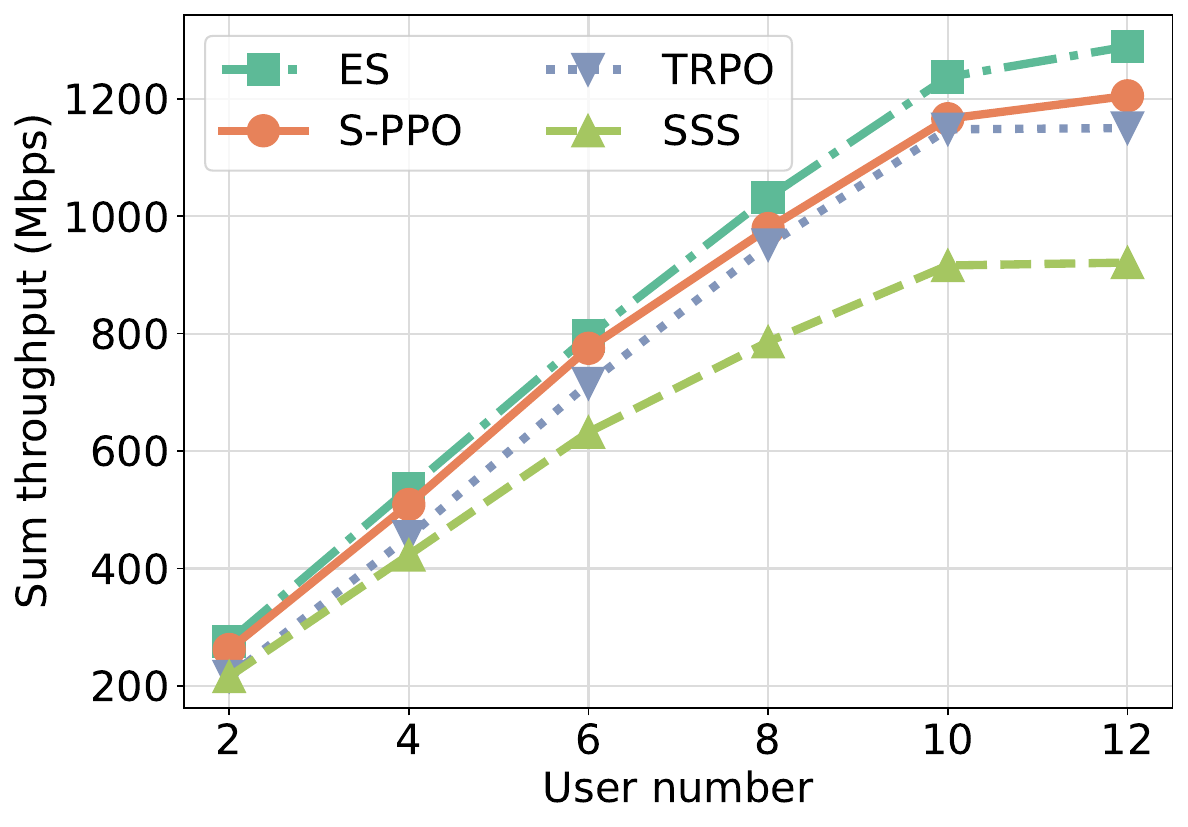}}\subfigure[Network setting 2 with 8 users]{
\label{fig:block_2}
\includegraphics[width=4.3cm,height = 3cm
]{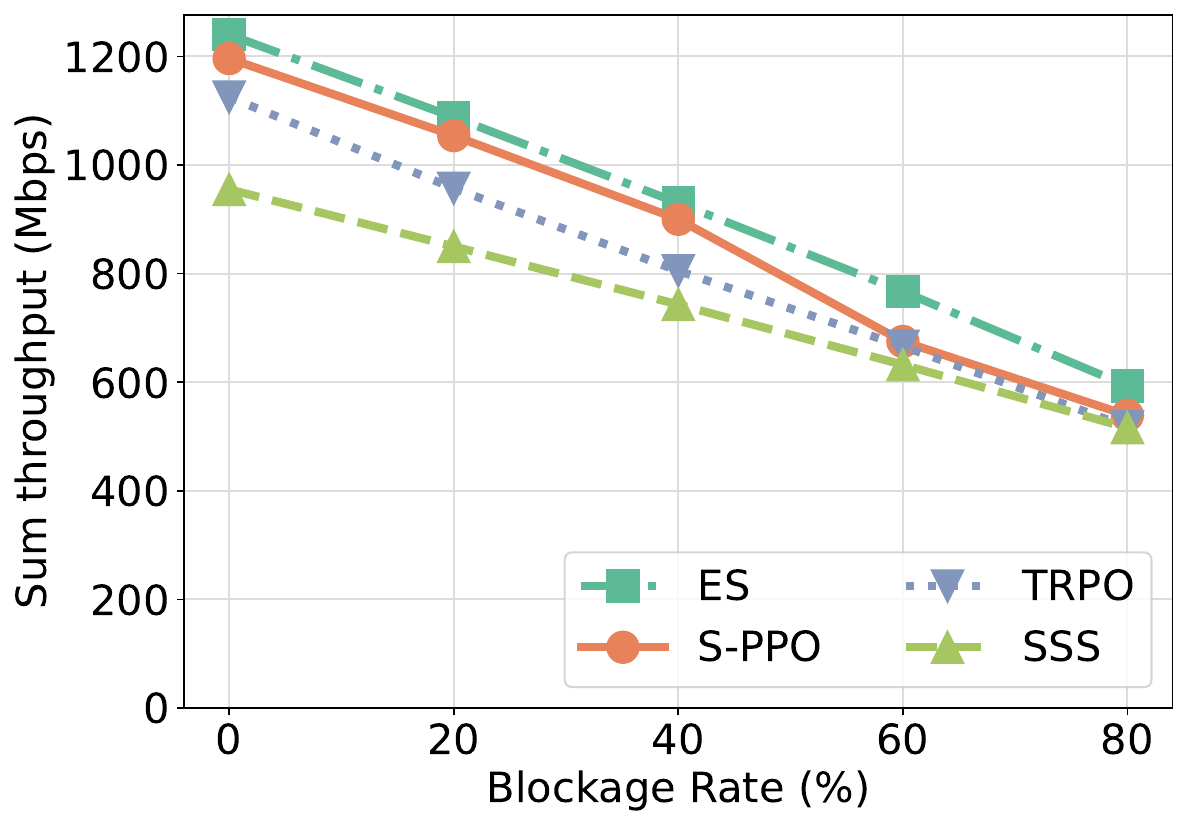}}
\caption{\small{Robustness comparison in terms of blockage rate in interference-prone scenario.}}
\label{fig:block}
\end{figure}

Considering the vulnerability of the VLC APs to blockage caused by obstacles, we next examine the robustness of the proposed S-PPO method in terms of blockage. Specifically, at each time slot, each user is randomly assigned a blockage probability to indicate whether blockage occurs. When the user’s random blockage probability is below the environment’s predefined blockage rate, blockage occurs, resulting in a drastically reduced SNR or SINR for the corresponding VLC AP. It is important to note that blockage only occurs when the VLC AP and the user are within each other’s field of view. Figure \ref{fig:block_1} shows the results under a 20\% blockage rate in the interference-prone scenario with network setting 1, where the user number ranges from 2 to 12. It can be observed that as the user number increases, S-PPO achieves near-optimal results and consistently outperforms TRPO and SSS. 
As expected, we also observe a slight reduction of sum throughput compared to the results in the same scenario but without blockage (Fig. \ref{fig:adap1}).
While Fig. \ref{fig:block_2} presents the sum throughput results for 8 users in the network setting 2 with respect to different blockage rates. Though the sum throughput decreases as the blockage rate increases, the proposed S-PPO method consistently achieves performance near the global optimum obtained by ES.

\begin{figure}[t]
\centering 
\vspace{-0.70mm}
\subfigure[Network setting 1]{
\label{fig:scal1}
\includegraphics[width=4.3cm,height = 3cm
]{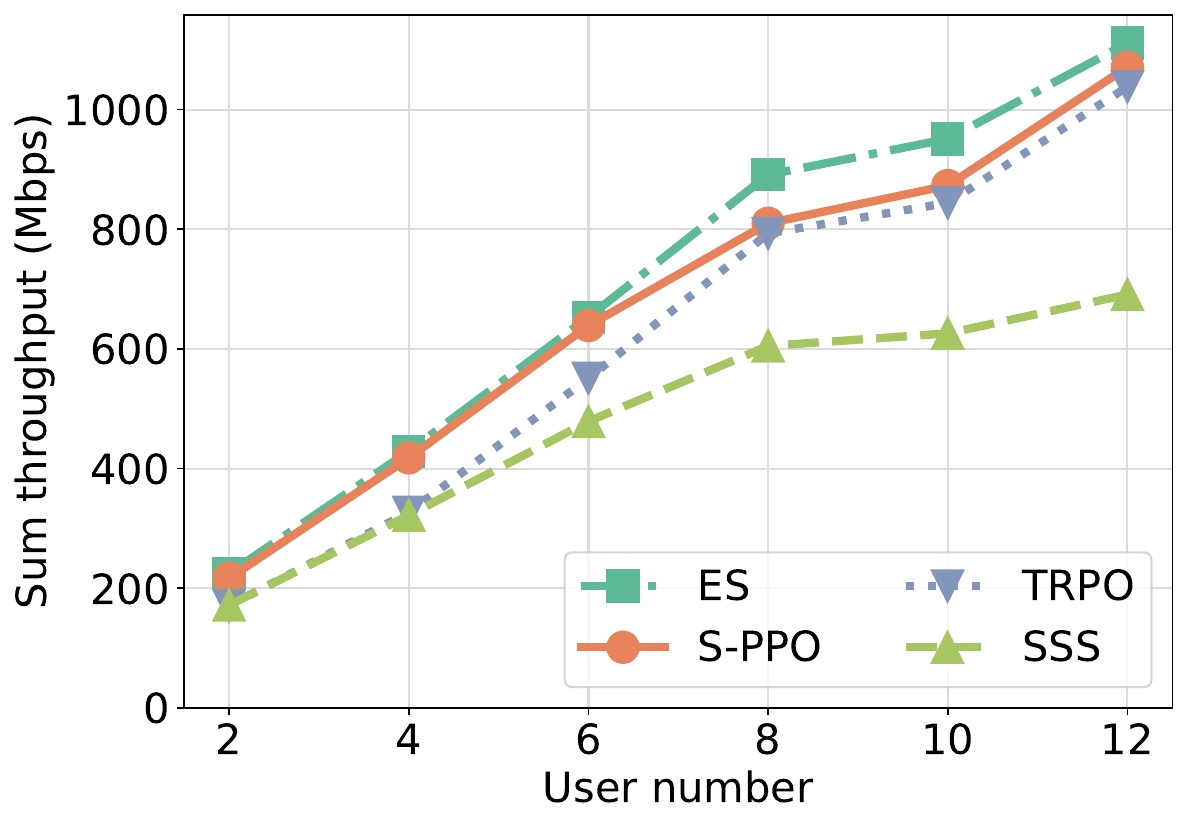}}\subfigure[Network setting 2]{
\label{fig:scal2}
\includegraphics[width=4.3cm,height = 3cm
]{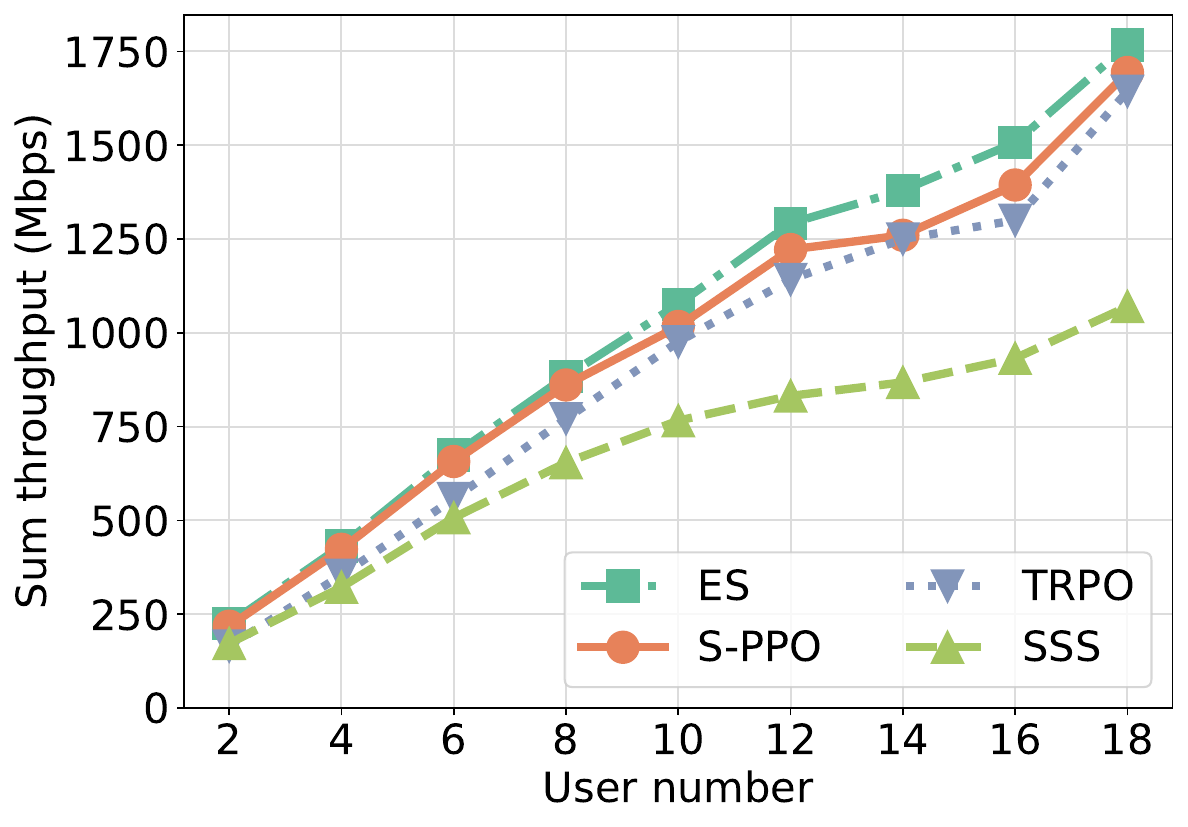}}
\caption{\small{Scalability comparison in interference-free scenario.}}
\vspace{-6mm}
\label{fig:scal_1}
\end{figure}

\begin{figure}[t]
\centering 
\subfigure[Network setting 1]{
\label{fig:adap1}
\includegraphics[width=4.3cm,height = 3cm
]{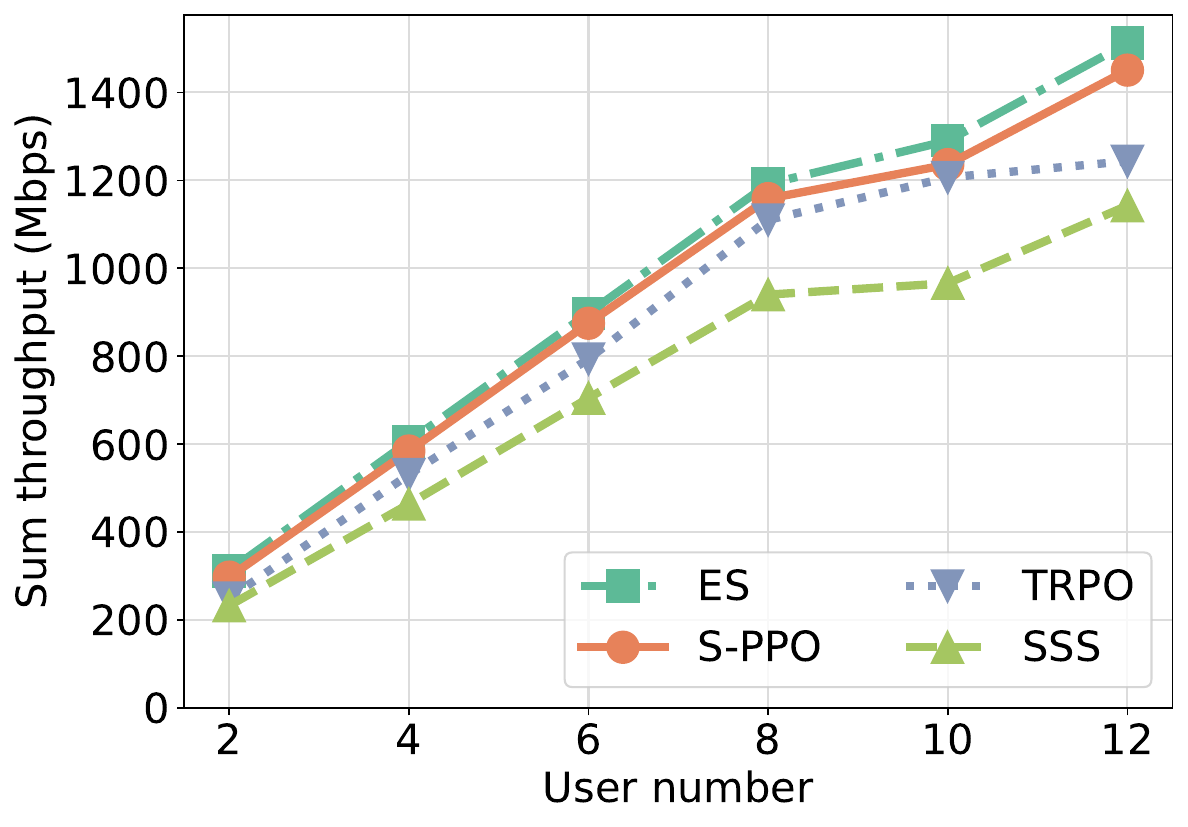}}\subfigure[Network setting 2]{
\label{fig:adap2}
\includegraphics[width=4.3cm,height = 3cm
]{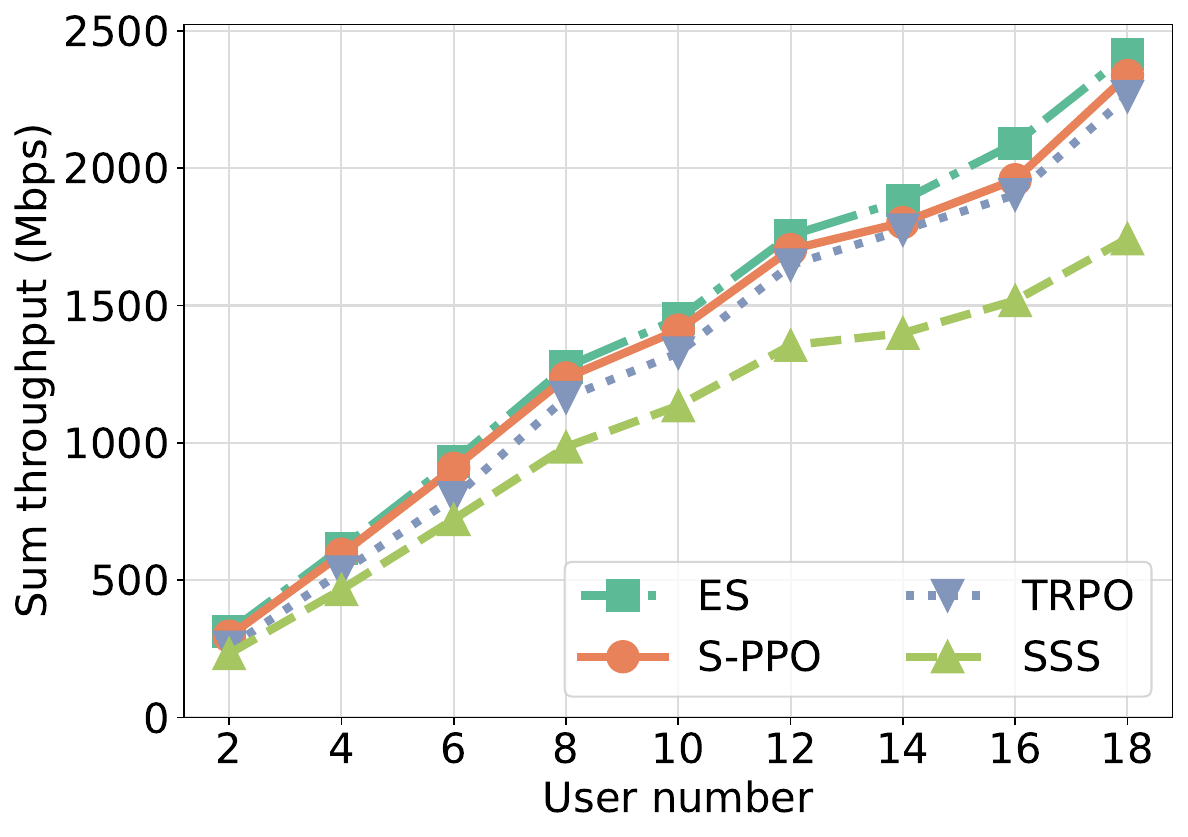}}
\caption{\small{Scalability comparison in interference-prone scenario.}}
\label{fig:adap_1}
\end{figure}
\begin{table}[b]
\centering
\vspace{-5mm}
\caption{Training time comparison in interference-prone scenario, network setting 2}
\label{table:runtime}
\begin{tabular}{|c|c|c|}
\hline
User number & S-PPO (s) & TRPO (s) \\ \hline
2 & \textbf{17.9} & 19.4 \\ \hline
4 & \textbf{39.6} & 43.6 \\ \hline
6 & \textbf{65.9} & 70.3 \\ \hline
8 & \textbf{94.0} & 102.3 \\ \hline
10 & \textbf{128.8} & 139.1 \\ \hline
12 & \textbf{172.2} & 184.6 \\ \hline
14 & \textbf{211.0} & 230.6 \\ \hline
16 & \textbf{249.9} & 276.9 \\ \hline
18 & \textbf{303.3} & 322.8 \\ \hline
\end{tabular}
\end{table}
\begin{figure}[t]
\centering 
\vspace{-2.5mm}
\subfigure[Dense network scenario topology]{
\label{fig:adap3}
\includegraphics[width=4.2cm,
]{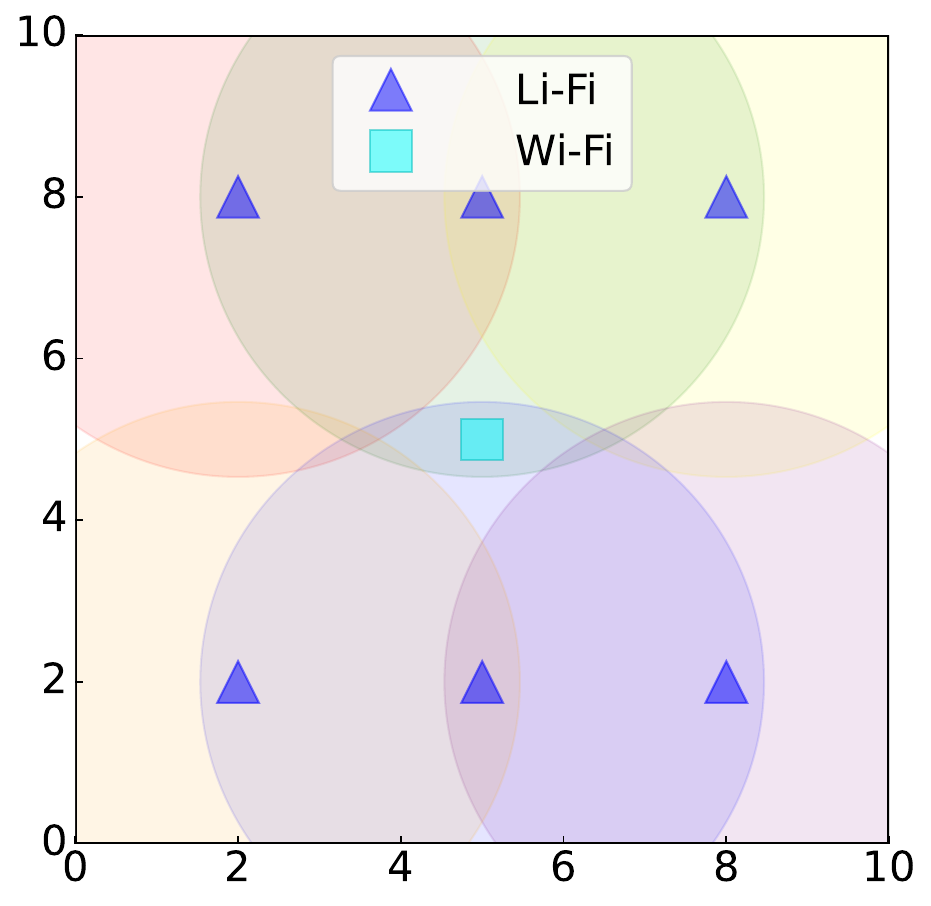}}
\subfigure[Network setting 1]{
\label{fig:adap4}
\includegraphics[width=4.3cm,height = 3cm
]{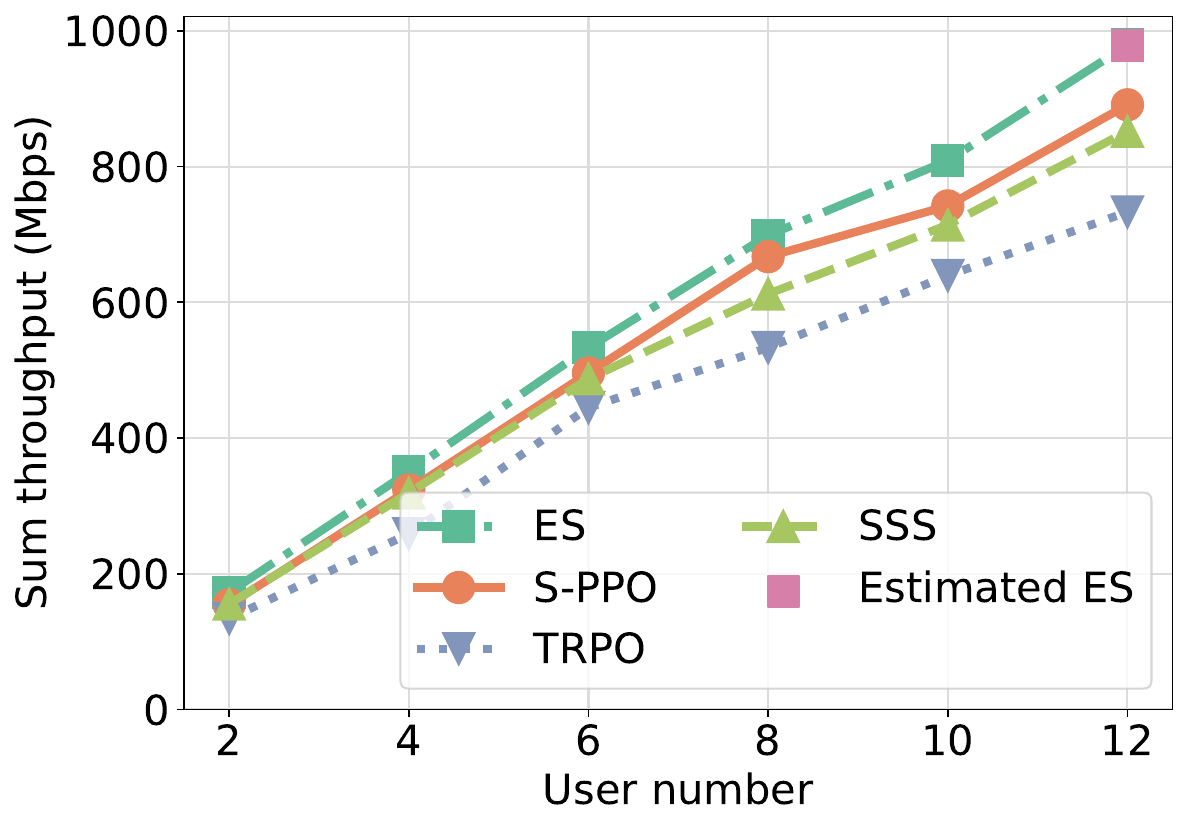}}
\caption{\small{Scalability comparison in dense network scenario.}}
\vspace{-4.5mm}
\label{fig:adap_2}
\end{figure}

\textbf{Computational complexity}: We further evaluate the computational complexity pertaining to the training time of the proposed S-PPO method by comparing it with that of TRPO. Specifically, we use a 2023 MacBook Pro with an Apple M2 Max chip, 32 GB of RAM, and a 1 TB SSD, operating under macOS Sonoma version 14.3. Python version 3.11.4 and PyTorch version 2.0.1 are used to implement the algorithms. The runtime comparison for the interference-prone scenario with network setting 2 is shown in Table \ref{table:runtime}. We can see that S-PPO consumes less time than TRPO across varying user numbers. It is important to note that even at a high user number of 18, the runtime of S-PPO is still within a feasible and acceptable range. This observation further indicates that the training time of S-PPO is relatively short, thus enabling its wide applicability in different network application scenarios. We also evaluate the inference time of well-trained S-PPO and TRPO in the 18-user 5-AP scenario. The results show that each inference run requires only about 0.000532 seconds and 0.000589 seconds, respectively. Thus, it demonstrates the practicality of S-PPO for real-time applications.

\textbf{Scalability}: 
To evaluate the scalability of the proposed S-PPO method, we then conduct extensive simulations by changing the user number from 2 to 18 and the AP number from 5 to 7 in different network settings. Figure \ref{fig:scal_1} shows the scalability comparison in two network settings under the interference-free scenario. We can obviously observe that as the number of users increases, the sum data rate increases. Moreover, compared with network setting 1 (Fig. \ref{fig:scal1}), network setting 2 (Fig. \ref{fig:scal2}) results in a higher sum throughput, particularly as the user number increases. This phenomenon is due to the increased capacity limitation in the network setting 2, where more users can be accommodated. Moreover, S-PPO constantly outperforms TRPO and SSS in both network settings, thus validating its scalability.

Figure \ref{fig:adap_1} illustrates the scalability results in the interference-prone scenario with respect to different user numbers. We can see that the proposed S-PPO method achieves nearly optimal results and outperforms TRPO and SSS. Furthermore, compared to the results of the interference-free scenario in Fig. 10, the achievable sum throughput in the interference-prone scenario is higher in both network setting 1 and network setting 2. This is because the coverage area of the LiFi APs is larger in the interference-prone scenario, even with potentially increased LiFi inter-cell interference, a higher throughput can still be provided by wider coverage of LiFi APs.

We further explore the scalability in a denser network scenario, where the number of LiFi APs increases to 6, as illustrated in Fig. \ref{fig:adap3}. The sum throughput comparison results in network setting 1 are shown in Fig. \ref{fig:adap4}, where we can see that S-PPO can achieve 92.2\% of the global optimal obtained by ES. The sum data rates denoted by pink markers are estimated values derived from the ES method because the computer cannot examine all the possible solutions within polynomial time. In this scenario, the sum throughput is not increased as expected. In contrast, the sum throughput decreases in all methods compared with that in the above-discussed 4-LiFi-AP interference-free and interference-prone scenarios. This is because the denser deployment of LiFi APs significantly increases inter-cell interference, which adversely impacts the throughput provided by LiFi APs, thus leading to a lower sum throughput. However, the proposed S-PPO algorithm can still achieve better performance in terms of sum throughput than both TRPO and SSS.

In conclusion, the proposed S-PPO method can achieve nearly optimal performance with the increase of user number and AP number in different network settings and can outperform both TRPO and SSS.

\begin{figure}[t]
\centering 
\subfigure[Network setting 1]{
\label{fig:fn1}
\includegraphics[width=4.3cm,height = 3cm
]{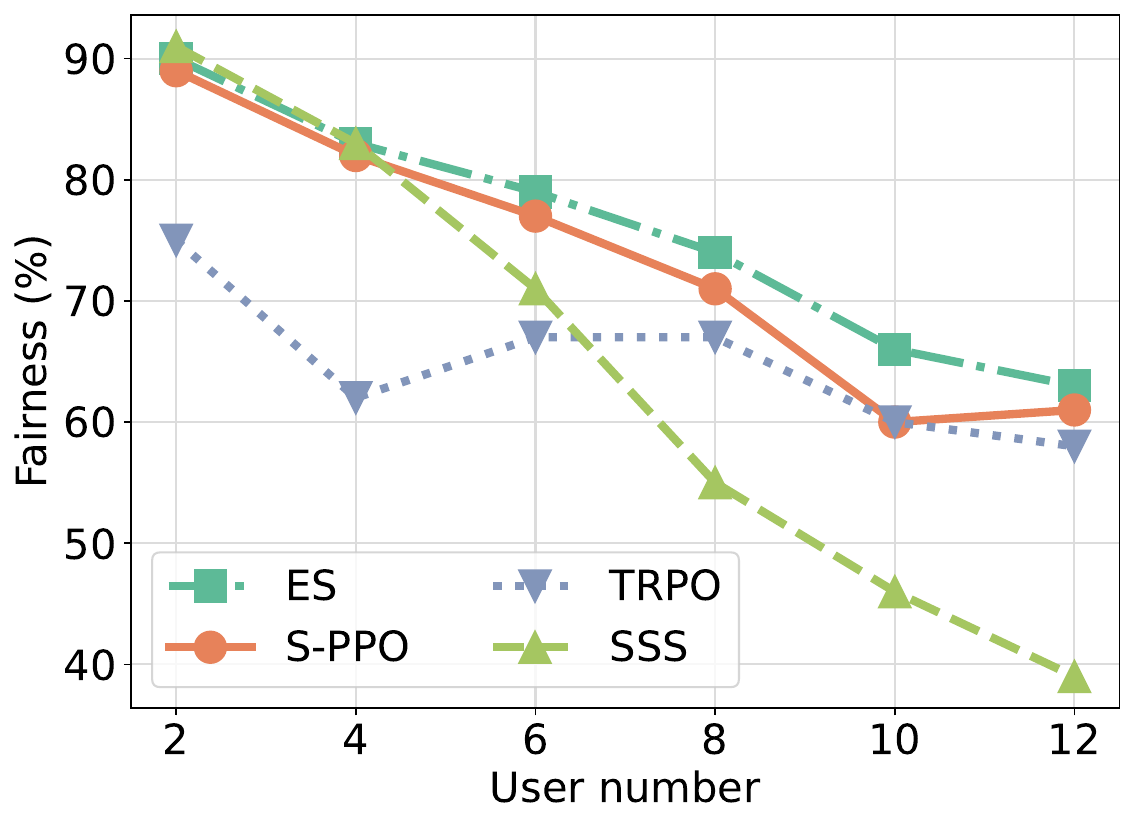}}\subfigure[Network setting 2]{
\label{fig:fn2}
\includegraphics[width=4.3cm,height = 3cm
]{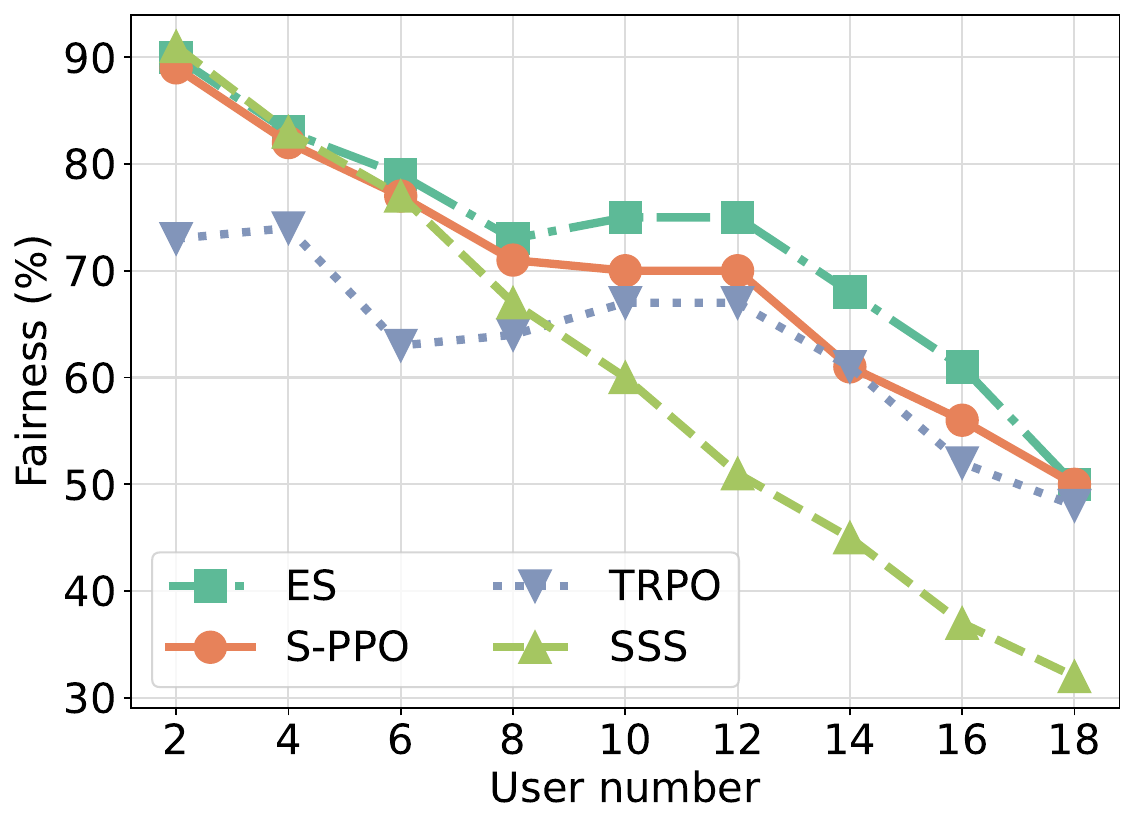}}
\caption{\small{Fairness comparison in interference-free scenario.}}
\vspace{-4mm}
\label{fig:fn_1}
\end{figure}

\textbf{Fairness}: Besides the sum throughput of the hybrid LiFi/WiFi networks, we also evaluate the fairness of the network, where Jain's fairness index \cite{jain1984quantitative} is adopted, given as:
\begin{figure}[t]
\centering 
\subfigure[Network setting 1]{
\label{fig:fn3}
\includegraphics[width=4.3cm,height = 3cm
]{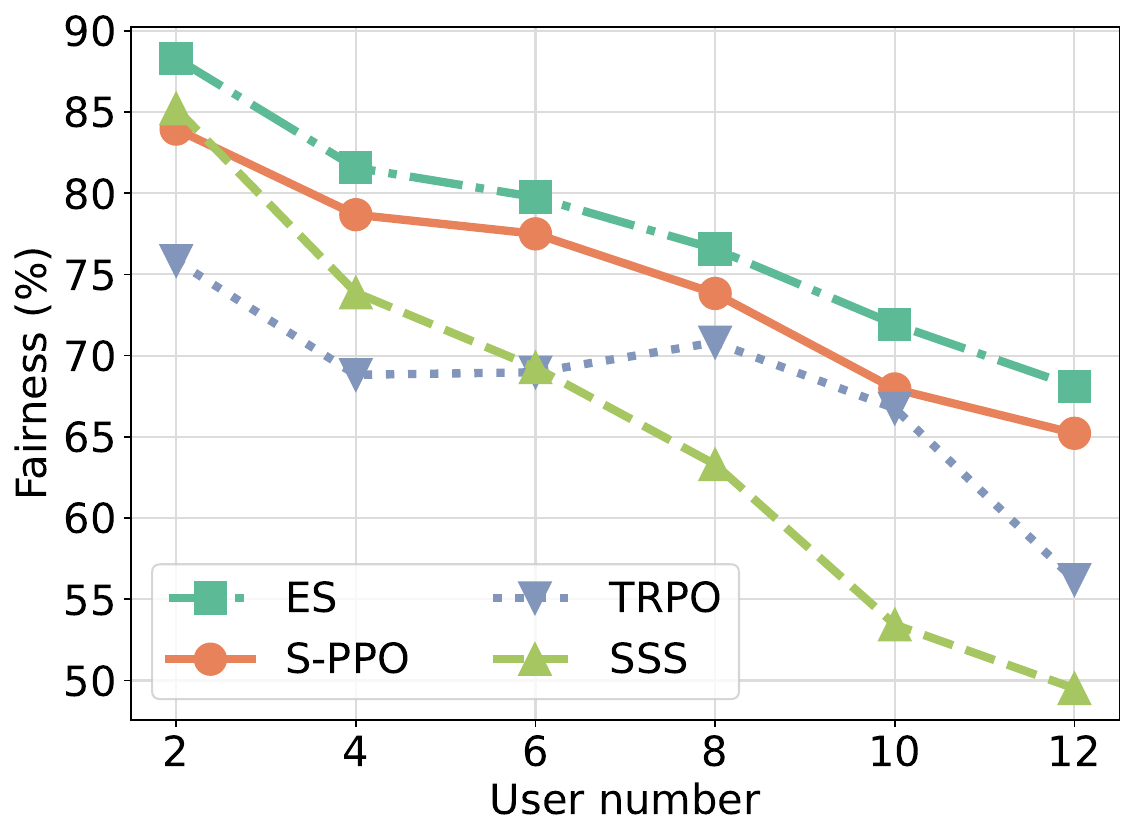}}\subfigure[Network setting 2]{
\label{fig:fn4}
\includegraphics[width=4.3cm,height = 3cm
]{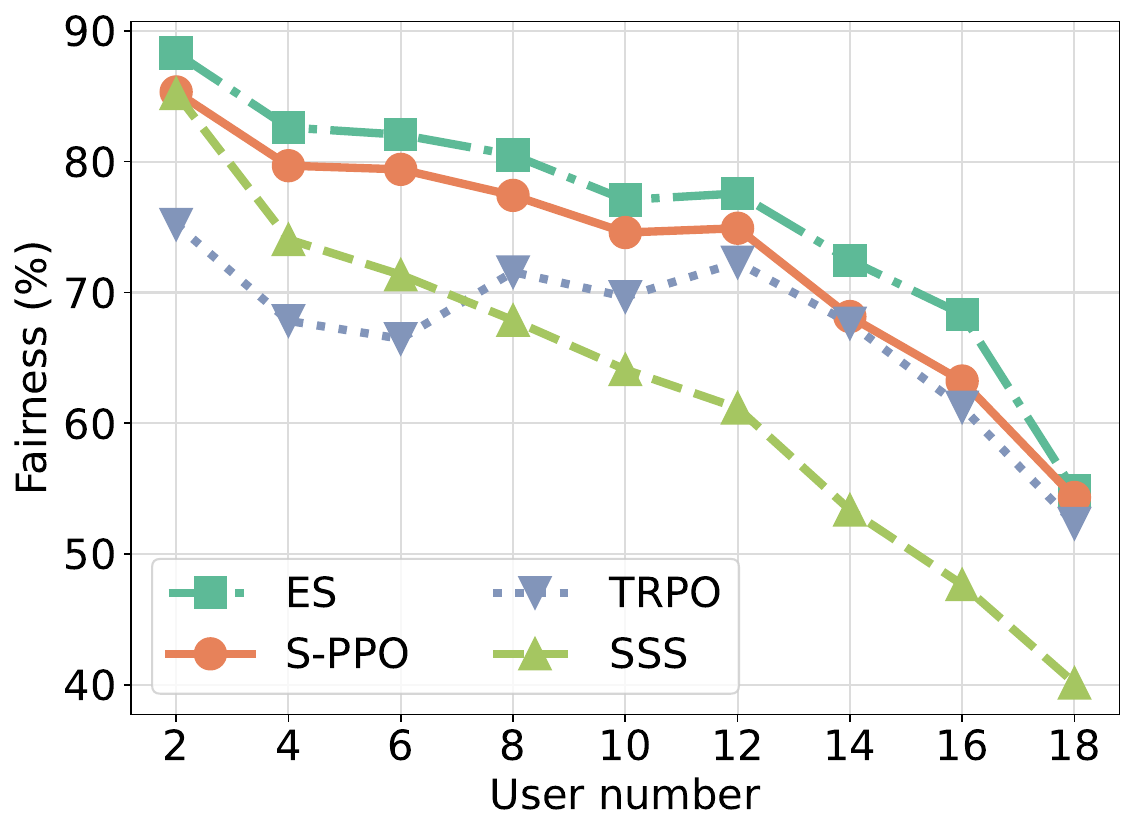}}
\caption{\small{Fairness comparison in interference-prone scenario.}}
\vspace{-4mm}
\label{fig:fn_2}
\end{figure}
\begin{align} \label{eq:fair}
\xi=\frac{(\sum\limits_{k=1}^{\mathcal{K}}r_k/r_k^{req})^2}{K\sum\limits_{k=1}^{\mathcal{K}}(r_k/r_k^{req})^2},
\end{align}where $r^{req}_{k}$ denotes the required data rate for user $k$. The required data rate is modeled as a random variable following a Gaussian distribution \cite{wu2017access}.

Figure \ref{fig:fn_1} and Fig. \ref{fig:fn_2} show the results of fairness in the interference-free and interference-prone scenarios respectively. As the number of users increases, we can see the fairness decreases. This is caused by the practical capacity limitation in hybrid LiFi/WiFi networks, where the growing demand due to the increasing user number cannot be completely accommodated. Therefore, the satisfaction ratio ($r_k/r_k^{req}$) decreases, thus resulting in a decrease in fairness. In addition, S-PPO outperforms TRPO and SSS. Specifically, with only two users, SSS exhibits greater fairness than ES in Fig. \ref{fig:fn3}. This is because the objective of ES is to maximize sum throughput, which doesn't guarantee optimal fairness. Therefore, SSS can obtain slightly better fairness than ES in this case. However, as the number of users increases, the performance of SSS drops dramatically and is worse than the other three methods, because it cannot satisfy all the user required data rates due to limited capacity.

To further evaluate the optimality of the fairness that the proposed S-PPO can achieve, we modify the objective function only related to fairness as:
\begin{align} \label{eq:ob_f}
    &\underset{\textbf{u}(t)}{\text{Maximize}} \quad f = \xi(t) \\
&\text{Subject to:}\;\;\;\ref{cs:locationx}, \ref{cs:locationy}, \ref{cs:lfbinary}, \ref{cs:wfbinary}, \ref{cs:nooverlap}, \ref{cs:lf}, \ref{cs:wf}.
\end{align}

\begin{figure}[t]
\centering 
\subfigure[Fairness]{
\label{fig:fn5}
\includegraphics[width=4.3cm,height = 3cm
]{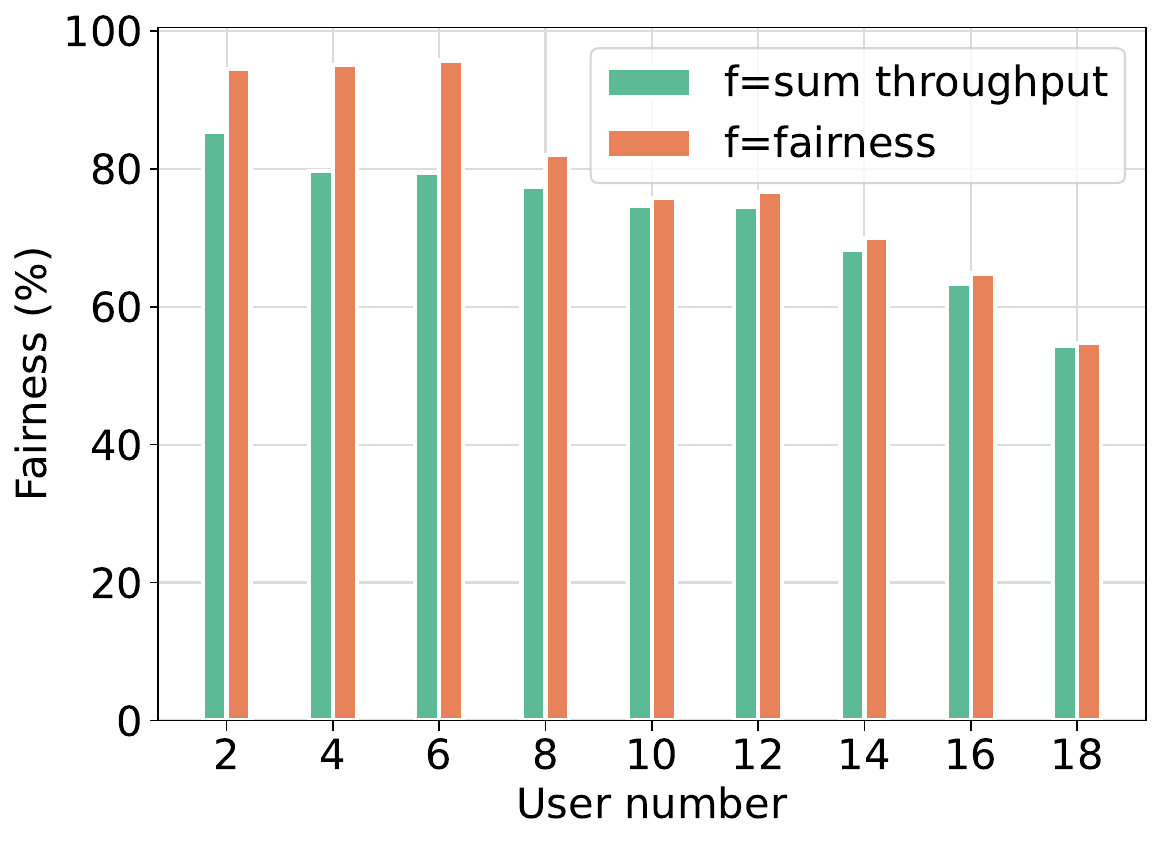}}\subfigure[Sum throughput]{
\label{fig:fn6}
\includegraphics[width=4.3cm,height = 3cm
]{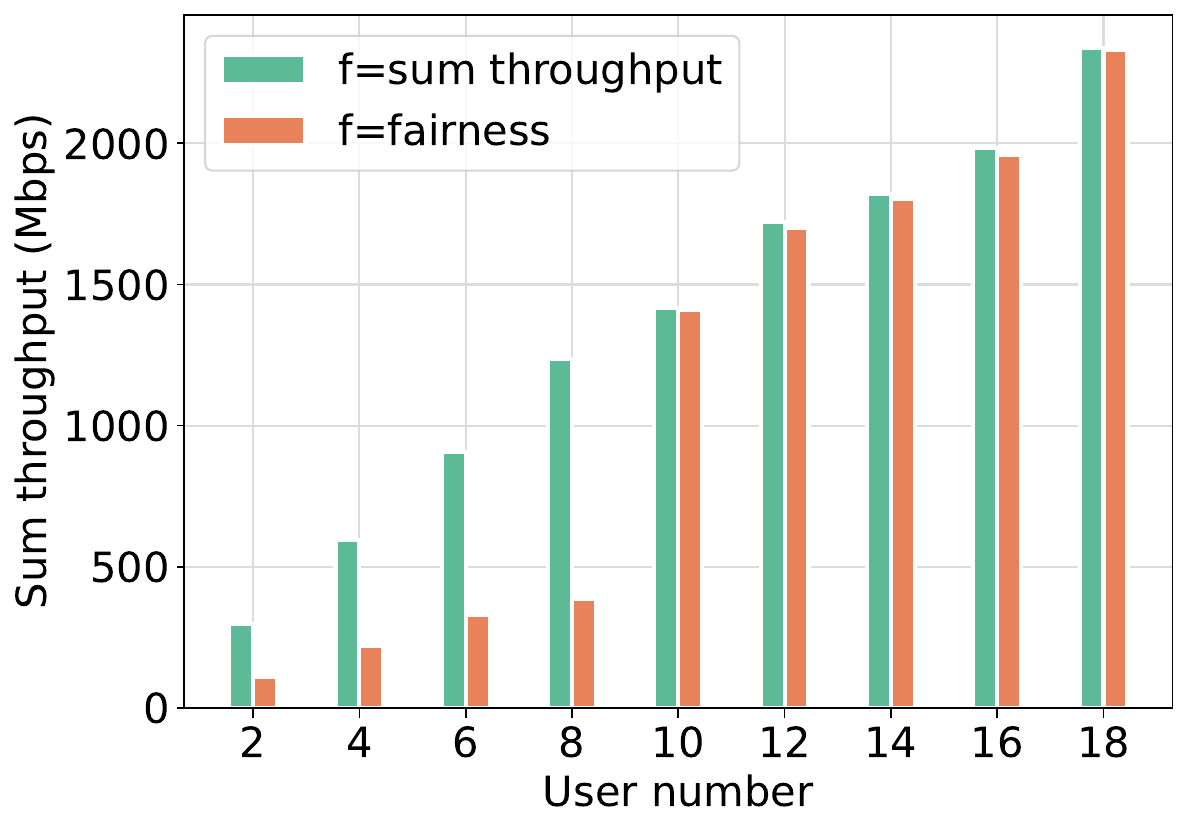}}
\caption{\small{Fairness-oriented optimization in interference-prone scenario, network setting 2.}}
\vspace{-6mm}
\label{fig:fairness}
\end{figure}

As depicted in Fig. \ref{fig:fn5}, the proposed S-PPO method can achieve higher fairness but significantly lower sum throughput compared with that achieved by the proposed sum throughput optimization problem when the number of users is less than 8. This may be due to the required user data rate may vary greatly when the user number is small. Therefore, maximizing fairness may sacrifice some user's data rate, thus resulting in a lower sum throughput than what is achieved by focusing solely on maximizing throughput without considering fairness. As the user number increases and their data rate requirements become more uniform, a fairness-focused approach can achieve throughput similar to those obtained when the objective function is sum throughput, even though the results for fairness do not demonstrate significant improvement. 

This phenomenon unveils the complex relationship between achieving fair resource allocation and maximizing overall network throughput. Relying solely on fairness as the objective function leads to inconsistent performance results, whereas our formulated objective function demonstrates both robust and consistent performance across various scenarios. In our future work, we plan to incorporate both fairness and sum throughput into our objective function to leverage the trade-off between them.

\section{Discussion}\label{sec:discussion}
\noindent \textbf{Deployment and Real-Time Implementation.} This work has demonstrated that the inference time of the proposed S-PPO achieves millisecond-level performance, enabling real-time user-AP association in dynamic environments. Additionally, this algorithm can be deployed on a central unit to gather network information and manage user-AP assignments. Therefore, this method is practical for real-world applications.

\noindent \textbf{Fairness in the Objective Function.} As mentioned in the previous section, while our work primarily focuses on optimizing the network sum throughput, other QoS metrics such as fairness are also essential for network performance. In future work, we plan to incorporate these QoS metrics into a multi-objective function. By balancing throughput, fairness, and other QoS factors, we aim to provide better user satisfaction and achieve load balancing across the hybrid networks.

\noindent \textbf{Scalability to Large-Scale Networks.} In this paper, we have conducted extensive simulations to evaluate the scalability of the proposed S-PPO under different scenarios with up to 7 APs and 18 users. Due to the limitations of devices in our lab, large-scale networks cannot be evaluated. However, distributed algorithms \cite{bayhan2020user} and graph neural networks \cite{deng2023gnn, zhang2023gnn} are promising approaches for addressing large-scale scenarios and we will leave it for future work.

\noindent \textbf{Adaptability to Varying AP and User Numbers.} In this work, we mainly focus on the scenarios with fixed AP and user numbers. However, in practical applications, the number of APs and users may vary over time, such as when users leave the network or new users join. Therefore, an adaptive method is needed for the varying environment. 
In future work, graph neural networks could be explored as a potential approach to model and address the challenges posed by varying AP and user numbers.

\vspace{-3mm}
\section{Conclusions}\label{sec:conclusion}

We study and formulate the user-AP association problem in hybrid indoor LiFi/WiFi networks mathematically, where the network control objective is to maximize the network-wide throughput by associating users to appropriate APs with consideration of user indoor mobility patterns and network capacity limitations. The resulting network control problem is a binary integer programming problem. We then propose the S-PPO method with the action space decomposition approach. Finally, we evaluate the proposed S-PPO method with extensive simulation results, comparing it with ES, SSS, and TRPO methods. Results show that the proposed solution algorithm can achieve $100$\% of the global optimum in static user scenario, and converge 267\% faster than TRPO in mobile user scenario. We also validate the optimality, convergence, robustness, computational complexity, scalability, and fairness of S-PPO in various scenarios. In the future, we plan to investigate the distributed algorithm to improve the scalability to large-scale networks and the adaptability to varying AP and user numbers. We will explore multi-objective optimization problem to leverage the tradeoff between different QoS metrics to further improve performance in hybrid LiFi/WiFi networks.

\bibliographystyle{ieeetr}
\bibliography{PPO/ppo}

\begin{thebibliography}{10}
\providecommand{\url}[1]{#1}
\csname url@samestyle\endcsname
\providecommand{\newblock}{\relax}
\providecommand{\bibinfo}[2]{#2}
\providecommand{\BIBentrySTDinterwordspacing}{\spaceskip=0pt\relax}
\providecommand{\BIBentryALTinterwordstretchfactor}{4}
\providecommand{\BIBentryALTinterwordspacing}{\spaceskip=\fontdimen2\font plus
\BIBentryALTinterwordstretchfactor\fontdimen3\font minus \fontdimen4\font\relax}
\providecommand{\BIBforeignlanguage}[2]{{%
\expandafter\ifx\csname l@#1\endcsname\relax
\typeout{** WARNING: IEEEtran.bst: No hyphenation pattern has been}%
\typeout{** loaded for the language `#1'. Using the pattern for}%
\typeout{** the default language instead.}%
\else
\language=\csname l@#1\endcsname
\fi
#2}}
\providecommand{\BIBdecl}{\relax}
\BIBdecl

\bibitem{ScreenTimeStats2023}
\BIBentryALTinterwordspacing
J.~Howarth. (2023) {Alarming Average Screen Time Statistics (2024)}. [Online]. Available: \url{https://explodingtopics.com/blog/screen-time-stats}
\BIBentrySTDinterwordspacing

\bibitem{komine2004fundamental}
T.~Komine and M.~Nakagawa, ``{Fundamental Analysis for Visible-Light Communication System Using LED Lights},'' \emph{IEEE transactions on Consumer Electronics}, vol.~50, no.~1, pp. 100--107, 2004.

\bibitem{cen2019lanet}
N.~Cen, J.~Jagannath, S.~Moretti, Z.~Guan, and T.~Melodia, ``{LANET: Visible-light Ad Hoc Networks},'' \emph{Ad Hoc Networks}, vol.~84, pp. 107--123, 2019.

\bibitem{my}
P.~Hou and N.~Cen, ``{Proximal Policy Optimization for User Association in Hybrid LiFi/WiFi Indoor Networks},'' in \emph{Proc. of IEEE GLOBECOM}, Kuala Lumpur, Malaysia, December 2023.

\bibitem{10060808}
Y.~Zhang and N.~Cen, ``{Programmable Software-Defined Testbed for Visible Light UAV Networks: Architecture Design and Implementation},'' in \emph{2023 IEEE 20th Consumer Communications \& Networking Conference (CCNC)}, Las Vegas, USA, January 2023, pp. 843--848.

\bibitem{10225783}
Y.~Long and N.~Cen, ``{Q-Learning for Sum-Throughput Optimization in Wireless Visible-Light UAV Networks},'' in \emph{IEEE INFOCOM 2023 - IEEE Conference on Computer Communications Workshops (INFOCOM WKSHPS)}, New York, USA, May 2023, pp. 1--6.

\bibitem{8574917}
P.~Pesek, S.~Zvánovec, P.~Chvojka, Z.~Ghassemlooy, and P.~A. Haigh, ``{Demonstration of a Hybrid FSO/VLC Link for the Last Mile and Last Meter Networks},'' \emph{IEEE Photonics Journal}, vol.~11, no.~1, pp. 1--7, 2019.

\bibitem{6967750}
A.~Gomez, K.~Shi, C.~Quintana, M.~Sato, G.~Faulkner, B.~C. Thomsen, and D.~O’Brien, ``{Beyond 100-Gb/s Indoor Wide Field-of-View Optical Wireless Communications},'' \emph{IEEE Photonics Technology Letters}, vol.~27, no.~4, pp. 367--370, 2015.

\bibitem{9351549}
X.~Wu, M.~D. Soltani, L.~Zhou, M.~Safari, and H.~Haas, ``{Hybrid LiFi and WiFi Networks: A Survey},'' \emph{IEEE Communications Surveys \& Tutorials}, vol.~23, no.~2, pp. 1398--1420, 2021.

\bibitem{rahaim2011hybrid}
M.~B. Rahaim, A.~M. Vegni, and T.~D. Little, ``{A Hybrid Radio Frequency and Broadcast Visible Light Communication System},'' in \emph{Proc. of IEEE GLOBECOM Workshops (GC Wkshps)}, Houston, USA, October 2011.

\bibitem{8926487}
J.~Kong, Z.-Y. Wu, M.~Ismail, E.~Serpedin, and K.~A. Qaraqe, ``{Q-Learning Based Two-Timescale Power Allocation for Multi-Homing Hybrid RF/VLC Networks},'' \emph{IEEE Wireless Communications Letters}, vol.~9, no.~4, pp. 443--447, 2020.

\bibitem{8792078}
H.~Yang, A.~Alphones, W.-D. Zhong, C.~Chen, and X.~Xie, ``{Learning-Based Energy-Efficient Resource Management by Heterogeneous RF/VLC for Ultra-Reliable Low-Latency Industrial IoT Networks},'' \emph{IEEE Transactions on Industrial Informatics}, vol.~16, no.~8, pp. 5565--5576, 2020.

\bibitem{9424627}
J.~Chen and Z.~Wang, ``{Coordination Game Theory-Based Adaptive Topology Control for Hybrid VLC/RF VANET},'' \emph{IEEE Transactions on Communications}, vol.~69, no.~8, pp. 5312--5324, 2021.

\bibitem{7510823}
Y.~Wang, X.~Wu, and H.~Haas, ``{Fuzzy logic based dynamic handover scheme for indoor Li-Fi and RF hybrid network},'' in \emph{2016 IEEE International Conference on Communications (ICC)}, Kuala Lumpur, Malaysia, May 2016, pp. 1--6.

\bibitem{pavon2003link}
J.~P. Pavon and S.~Choi, ``{Link Adaptation Strategy for IEEE 802.11 WLAN via Received Signal Strength Measurement},'' in \emph{Proc. of IEEE International Conference on Communications}, vol.~2, Anchorage, USA, May 2003.

\bibitem{basnayaka2017design}
D.~A. Basnayaka and H.~Haas, ``{Design and Analysis of a Hybrid Radio Frequency and Visible Light Communication System},'' \emph{IEEE Transactions on Communications}, vol.~65, no.~10, pp. 4334--4347, 2017.

\bibitem{wu2019mobility}
X.~Wu and H.~Haas, ``{Mobility-Aware Load Balancing for Hybrid LiFi and WiFi Networks},'' \emph{Journal of Optical Communications and Networking}, vol.~11, no.~12, pp. 588--597, 2019.

\bibitem{8123892}
J.~Wang, C.~Jiang, H.~Zhang, X.~Zhang, V.~C.~M. Leung, and L.~Hanzo, ``{Learning-Aided Network Association for Hybrid Indoor LiFi-WiFi Systems},'' \emph{IEEE Transactions on Vehicular Technology}, vol.~67, no.~4, pp. 3561--3574, 2018.

\bibitem{8374416}
Z.~Du, C.~Wang, Y.~Sun, and G.~Wu, ``{Context-Aware Indoor VLC/RF Heterogeneous Network Selection: Reinforcement Learning With Knowledge Transfer},'' \emph{IEEE Access}, vol.~6, pp. 33\,275--33\,284, 2018.

\bibitem{ahmad2020reinforcement}
R.~Ahmad, M.~D. Soltani, M.~Safari, A.~Srivastava, and A.~Das, ``{Reinforcement Learning Based Load Balancing for Hybrid LiFi WiFi Networks},'' \emph{IEEE Access}, vol.~8, pp. 132\,273--132\,284, 2020.

\bibitem{8943127}
X.~Wu and H.~Haas, ``{Load Balancing for Hybrid LiFi and WiFi Networks: To Tackle User Mobility and Light-Path Blockage},'' \emph{IEEE Transactions on Communications}, vol.~68, no.~3, pp. 1675--1683, 2020.

\bibitem{10661225}
I.~W.~G. da~Silva, E.~Eduardo Benitez~Olivo, M.~Katz, and D.~Pamela Moya~Osorio, ``{Analysis and Simulation of Precoding and User Association for Securing Hybrid RF/VLC Systems},'' \emph{IEEE Sensors Journal}, vol.~24, no.~20, pp. 33\,467--33\,480, 2024.

\bibitem{7876858}
Y.~Wang, X.~Wu, and H.~Haas, ``{Load Balancing Game With Shadowing Effect for Indoor Hybrid LiFi/RF Networks},'' \emph{IEEE Transactions on Wireless Communications}, vol.~16, no.~4, pp. 2366--2378, 2017.

\bibitem{wu2017access}
X.~Wu, M.~Safari, and H.~Haas, ``{Access Point Selection for Hybrid Li-Fi and Wi-Fi Networks},'' \emph{IEEE Transactions on Communications}, vol.~65, no.~12, pp. 5375--5385, 2017.

\bibitem{10367833}
H.~Ji, X.~Wu, Q.~Wang, S.~J. Redmond, and I.~Tavakkolnia, ``{Adaptive Target-Condition Neural Network: DNN-Aided Load Balancing for Hybrid LiFi and WiFi Networks},'' \emph{IEEE Transactions on Wireless Communications}, vol.~23, no.~7, pp. 7307--7318, 2024.

\bibitem{guerin2021towards}
E.~Gu{\'e}rin, T.~Begin, A.~Busson, and I.~Gu{\'e}rin~Lassous, ``{Towards a Throughput and Energy Efficient Association Strategy for Wi-Fi/LiFi Heterogeneous Networks},'' in \emph{Proceedings of the 18th ACM Symposium on Performance Evaluation of Wireless Ad Hoc, Sensor, \& Ubiquitous Networks}, Alicante, Spain, November 2021, pp. 119--126.

\bibitem{9468898}
R.~Ahmad, M.~D. Soltani, M.~Safari, and A.~Srivastava, ``{Reinforcement Learning-Based Near-Optimal Load Balancing for Heterogeneous LiFi WiFi Network},'' \emph{IEEE Systems Journal}, vol.~16, no.~2, pp. 3084--3095, 2022.

\bibitem{9127161}
M.~Sana, A.~De~Domenico, W.~Yu, Y.~Lostanlen, and E.~Calvanese~Strinati, ``{Multi-Agent Reinforcement Learning for Adaptive User Association in Dynamic mmWave Networks},'' \emph{IEEE Transactions on Wireless Communications}, vol.~19, no.~10, pp. 6520--6534, 2020.

\bibitem{kahn1997wireless}
J.~Kahn and J.~Barry, ``{Wireless Infrared Communications},'' \emph{Proceedings of the IEEE}, vol.~85, no.~2, pp. 265--298, 1997.

\bibitem{6636053}
J.-B. Wang, Q.-S. Hu, J.~Wang, M.~Chen, and J.-Y. Wang, ``{Tight Bounds on Channel Capacity for Dimmable Visible Light Communications},'' \emph{Journal of Lightwave Technology}, vol.~31, no.~23, pp. 3771--3779, 2013.

\bibitem{perahia2013next}
E.~Perahia and R.~Stacey, \emph{{Next Generation Wireless LANs: 802.11 n and 802.11 ac}}.\hskip 1em plus 0.5em minus 0.4em\relax Cambridge University Press, 2013.

\bibitem{johnson1996dynamic}
D.~B. Johnson and D.~A. Maltz, ``{Dynamic Source Routing in Ad Hoc Wireless Networks},'' in \emph{Mobile computing}.\hskip 1em plus 0.5em minus 0.4em\relax Springer, 1996, pp. 153--181.

\bibitem{boyd2004convex}
S.~Boyd, ``{Convex Optimization},'' \emph{Cambridge UP}, 2004.

\bibitem{konda1999actor}
V.~Konda and J.~Tsitsiklis, ``{Actor-Critic Algorithms},'' \emph{Advances in Neural Information Processing Systems}, vol.~12, pp. 1008--1024, 2000.

\bibitem{schulman2017proximal}
J.~Schulman, F.~Wolski, P.~Dhariwal, A.~Radford, and O.~Klimov, ``{Proximal Policy Optimization Algorithms},'' \emph{arXiv preprint arXiv:1707.06347}, 2017.

\bibitem{10228875}
X.~He, X.~Zhuge, F.~Dang, W.~Xu, and Z.~Yang, ``{DeepScheduler: Enabling Flow-Aware Scheduling in Time-Sensitive Networking},'' in \emph{IEEE INFOCOM 2023 - IEEE Conference on Computer Communications}, New York, USA, May 2023, pp. 1--10.

\bibitem{jain1984quantitative}
R.~K. Jain, D.-M.~W. Chiu, W.~R. Hawe \emph{et~al.}, ``{A Quantitative Measure of Fairness and Discrimination},'' \emph{Eastern Research Laboratory, Digital Equipment Corporation, Hudson, MA}, vol.~21, p.~1, 1984.

\bibitem{bayhan2020user}
S.~Bayhan, E.~Coronado, R.~Riggio, and A.~Zubow, ``{User-AP Association Management in Software-defined WLANs},'' \emph{IEEE Transactions on Network and Service Management}, vol.~17, no.~3, pp. 1838--1852, 2020.

\bibitem{deng2023gnn}
W.~Deng, Y.~Liu, M.~Li, and M.~Lei, ``{GNN-Aided User Association and Beam Selection for mmWave Integrated Heterogeneous Networks},'' \emph{IEEE Wireless Communications Letters}, 2023.

\bibitem{zhang2023gnn}
H.~Zhang, X.~Ma, X.~Liu, L.~Li, and K.~Sun, ``{GNN-based Power Allocation and User Association in Digital Twin Network for the Terahertz Band},'' \emph{IEEE Journal on Selected Areas in Communications}, 2023.

\end{thebibliography}
\end{document}